\documentclass[10pt,twocolumn,letterpaper]{article}
\pdfoutput=1

\usepackage{iccv}
\usepackage{times}
\usepackage{epsfig}
\usepackage{graphicx}
\usepackage{amsmath}
\usepackage{amssymb}

\usepackage[square,numbers,sort&compress]{natbib}
\usepackage{enumitem}
\usepackage{algpseudocode}
\usepackage[dvipsnames]{xcolor}
\usepackage[font=small,labelfont=bf]{caption}
\usepackage{array}
\usepackage{multirow}
\usepackage{booktabs}
\usepackage{algorithm}
\usepackage[normalem]{ulem}
\usepackage{xparse}
\usepackage{pifont}
\usepackage{bm}
\usepackage{threeparttable}
\usepackage{adjustbox}

\usepackage{listings}
\usepackage{mwe} %
\usepackage{makecell}
\usepackage{color, colortbl}
\usepackage{footmisc}  %

\usepackage[accsupp]{axessibility}  %

\usepackage[breaklinks=true,bookmarks=false,colorlinks,bookmarks=false, citecolor=ForestGreen]{hyperref}
\usepackage{cleveref}

\iccvfinalcopy %

\def\iccvPaperID{1123} %

\ificcvfinal\fi

\crefname{section}{\S}{\S\S}
\crefname{subsection}{\S}{\S\S}
\crefformat{table}{Table~#2#1#3}
\crefformat{figure}{Fig.~#2#1#3}
\crefformat{equation}{Eq.~#2#1#3}

\newcommand{\calB}{\mathcal{B}}

\newcommand{\calL}{\mathcal{L}}
\newcommand{\calD}{\mathcal{D}}

\newcommand{\cmark}{\ding{51}}%
\newcommand{\eknewfull}{EpicKitchens-100\xspace}
\newcommand{\eknew}{EK100\xspace}
\newcommand{\ekfull}{EpicKitchens-55\xspace}
\newcommand{\ek}{EK55\xspace}
\newcommand{\saladfull}{50-Salads\xspace}
\newcommand{\salad}{50S\xspace}
\newcommand{\egtea}{EGTEA Gaze+\xspace}
\newcommand{\imnet}{IN1k\xspace}
\newcommand{\imnetii}{IN21k\xspace}
\newcommand{\imnetboth}{IN21+1k\xspace}

\newcommand{\rulstm}{RULSTM\xspace}
\newcommand{\fhoi}{FHOI\xspace}
\newcommand{\acbank}{ActionBanks\xspace}
\newcommand{\imrnn}{ImagineRNN\xspace}
\newcommand{\sota}{state-of-the-art\xspace}
\newcommand{\correctSample}{Default\xspace}
\newcommand{\methodfull}{Anticipative Video Transformer\xspace}
\newcommand{\method}{AVT\xspace}
\newcommand{\methodfused}{AVT+\xspace}
\newcommand{\methodfusedall}{AVT++\xspace}
\newcommand{\txBack}{\method{}-b\xspace}
\newcommand{\txHead}{\method{}-h\xspace}

\newcommand{\causalSetting}{anticipative\xspace}
\newcommand{\causalSettingShort}{[a]\xspace}
\newcommand{\acausalSetting}{naive\xspace}
\newcommand{\acausalSettingShort}{[n]\xspace}
\newcommand{\lossFuture}{\calL_{next}}
\newcommand{\lossPast}{\calL_{cls}}
\newcommand{\lossGPT}{\calL_{feat}}

\newcommand{\lossBoth}{anticipative\xspace}
\newcommand{\LossBoth}{Anticipative\xspace}

\newlength\savewidth

\newlength\thinwidth

\definecolor{Gray}{gray}{0.92}
\newcolumntype{g}{>{\columncolor{Gray}}c}
\definecolor{LightCyan}{rgb}{0.88,1,1}
\definecolor{altRowColor}{gray}{0.92}
\definecolor{highlightRowColor}{rgb}{0.95, 0.95, 1}

\definecolor{demphcolor}{RGB}{100,100,100}

\newcolumntype{H}{>{\setbox0=\hbox\bgroup}c<{\egroup}@{}}
\definecolor{highlightRowColor}{rgb}{0.95, 0.95, 1}

\crefname{section}{\S}{\S\S}
\crefname{subsection}{\S}{\S\S}
\crefformat{table}{Table~#2#1#3}
\crefformat{figure}{Figure~#2#1#3}
\crefformat{equation}{Equation~(#2#1#3)}
\crefformat{algorithm}{Algorithm~#2#1#3}
\crefformat{appendix}{Appendix~#2#1#3}

\newif\ifshortappdx

\begin{document}

\title{\methodfull}

\author{
Rohit Girdhar$^\dagger$ \qquad Kristen Grauman$^{\dagger\ddagger}$ \\
{$^\dagger$Facebook AI Research \quad $^\ddagger$University of Texas, Austin} \\
{\small \url{http://facebookresearch.github.io/AVT}}
}

\maketitle
\ificcvfinal\thispagestyle{empty}\fi

\begin{abstract}
    We propose \methodfull (\method), an end-to-end attention-based %
    video modeling architecture that
    attends to %
    the previously observed video in order to anticipate future actions.  We train the model jointly to predict the next action in a video sequence, while also learning frame feature encoders that are 
    predictive
    of successive future frames' features.  Compared to existing temporal aggregation strategies, \method has the advantage of both maintaining the sequential progression of observed actions while still capturing long-range dependencies---both critical for the anticipation task. Through extensive experiments, we show that
    \method obtains the best reported performance on four popular
    action anticipation benchmarks: \ekfull, \eknewfull, \egtea, and \saladfull; and it wins %
    first place
    in the \eknewfull CVPR'21 challenge.
\end{abstract}
\section{Introduction}

\begin{figure}[t]
    \centering
    \includegraphics[width=\linewidth,bb=0 0 541 275]{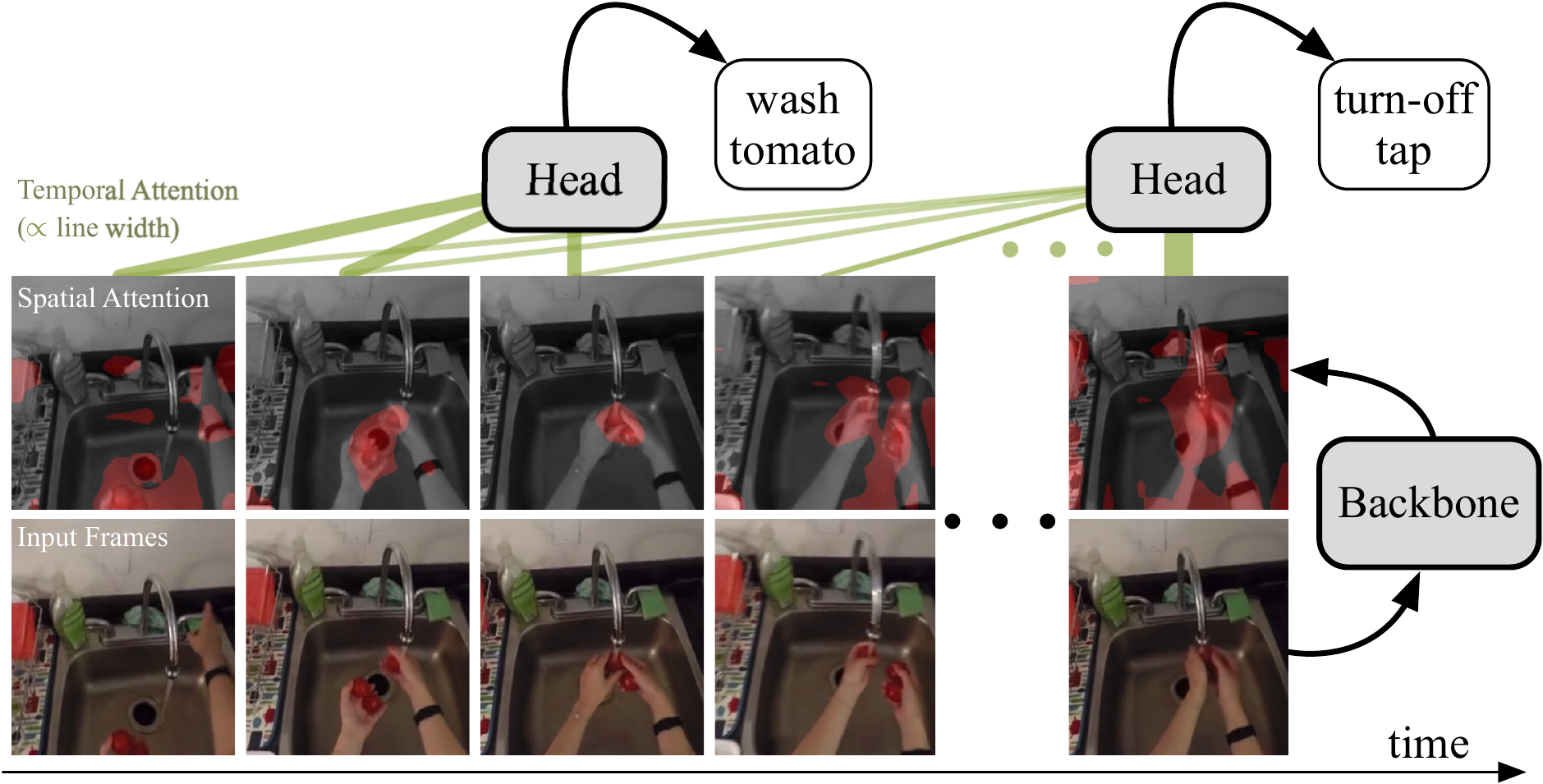}  %
    \vspace*{-0.25in}
    \caption{
        {\bf Anticipating future actions using \method}
        involves encoding video frames with a spatial-attention backbone, followed by 
        a temporal-attention head that attends only to frames before the current one to predict future actions.
        In this example, it spontaneously learns to attend to hands and objects without being supervised to do so. %
        Moreover, it attends to frames most relevant to predict the next action.
        For example, to predict `wash tomato' it attends equally to all previous frames as they %
        determine if any more tomatoes need to be washed, whereas for
        `turn-off tap' it focuses most on the current frame for cues %
        whether the person might be done. Please see~\cref{sec:expt:qual:att} for details and additional results.
    }
    \label{fig:teaser}
\end{figure}

Predicting future human actions is an important task for AI systems. Consider an autonomous vehicle at a stop sign that needs to predict whether a pedestrian will cross the street or not. Making this determination  %
requires modeling complex visual signals---the past actions of the pedestrian, such as speed and direction of walking, or usage of devices that may hinder 
his awareness of the surroundings---and using %
those to predict what he may do next. Similarly, imagine an augmented reality (AR) device that observes a user's activity from a wearable camera, \eg as they cook a new dish or assemble a piece of furniture, and needs to anticipate his next steps to provide timely assistance. %
In many such applications,
it is insufficient to \emph{recognize} what is happening in the video.  Rather, the vision system must also \emph{anticipate} the likely actions that are to follow.
Hence, there is a growing interest in formalizing the \emph{activity anticipation} task~\cite{kitani2012activity,stein2013combining,kuehne2014language,rhinehart2017first,furnari2020rulstm,nagrajan2020egotopo} 
along with development of multiple challenge benchmarks to support it~\cite{stein2013combining,Damen2018EPICKITCHENS,li2018eye,damen2020rescaling,kuehne2014language}.%

Compared to %
traditional action recognition, anticipation tends to be significantly more challenging.  First of all, 
it requires going beyond %
classifying current spatiotemporal visual patterns into a single action category---a task nicely suited to today's well-honed discriminative models---to instead predict the multi-modal distribution of future activities. %
Moreover, while action recognition can often side-step temporal reasoning by leveraging instantaneous contextual cues~\cite{girdhar2020cater}, anticipation %
inherently requires modeling the progression of past actions to predict the future. 
For instance, the presence of a plate of food with a fork may be sufficient to indicate the action of eating, whereas anticipating that same action would require recognizing and reasoning over the sequence of actions that precede it, such as chopping, cooking, serving, \etc.
Indeed, recent work~\cite{furnari2019rulstm,sener2020temporal} finds that modeling long temporal context is often critical for anticipation, %
unlike action recognition where frame-level modeling is often enough~\cite{ucf101,hmdb51,kay2017kinetics}. 
These challenges are also borne out in practice. For example, accuracy for one of today's  top performing video models~\cite{sener2020temporal} drops from 42\% to 17\% when treating recognition versus anticipation on the same test clips~\cite{Damen2018EPICKITCHENS}---predicting even one second into the future is much harder than declaring the current action.

The typical approach to solving long-term predictive reasoning tasks involves extracting frame or clip level features using standard architectures~\cite{wang2016tsn,tran2015learning,carreira2017quo}, followed by aggregation using clustering~\cite{Girdhar_17a_ActionVLAD,miech17loupe}, recurrence~\cite{furnari2019rulstm,furnari2020rulstm,Karpathy_14}, or attention~\cite{long2017attention,wu2019long,girdhar2019video,sener2020temporal} based models. Except the recurrent ones, most such models merely aggregate features over the temporal extent, with little regard to modeling the sequential temporal evolution of the video over frames. While recurrent models like LSTMs %
have been explored for anticipation~\cite{furnari2019rulstm,abu2018will,wu2021imaginernn}, they %
are known to struggle with modeling long-range temporal dependencies due to their sequential (non-parallel) nature.   
Recent work %
mitigates this limitation using attention-based aggregation over different amounts of the context to produce short-term (`recent') and long-term (`spanning') features~\cite{sener2020temporal}. However, it still reduces the video to multiple aggregate representations and loses its sequential nature. %
Moreover, it relies on careful and dataset-specific tuning of the architecture and the amounts of context used for the different aggregate features.

In this work, we introduce \emph{\methodfull} (\method), an alternate video modeling architecture that replaces ``aggregation'' based temporal modeling with a 
{\em \lossBoth{}}\footnote{We use the term ``\lossBoth{}'' to refer to our model's ability to predict future video features and actions.}
architecture.  
Aiming to overcome the tradeoffs described above, the proposed model naturally embraces the %
sequential nature of videos, while minimizing the limitations that arise with recurrent architectures. 
Similar to recurrent models, \method can be rolled out indefinitely to predict further into the future (\ie generate future predictions), yet it does so while processing the input in parallel with long-range attention, which is often lost in recurrent architectures.

Specifically, %
\method leverages the popular transformer architecture~\cite{vaswani2017attention,wang2019learning} 
with {\em causal}\footnote{Throughout we use the term ``causal'' to refer to the constraint that video be processed in a forward, online manner, \ie functions applied at time $t$ can only reference the frames preceding them, 
akin to Causal Language Modeling (CLM)~\cite{lample2019cross}.
This is not to be confused with other uses of ``causal'' in AI where the connotation is instead cause-and-effect.}
masked attention, %
where each input frame is allowed to attend only to frames that precede it. We train the model to jointly predict the next action while also learning to predict future features that match the true future features and (when available) their intermediate action labels.
\cref{fig:teaser} shows examples of how \method{}'s spatial and temporal attention spreads over previously observed frames for two of its future predictions (wash tomato and turn-off tap). 
By %
incorporating intermediate %
future prediction losses, \method encourages a %
predictive
video representation that picks up patterns in how the visual activity is likely to unfold into the future.
This facet of our model draws an analogy to language, where transformers trained with massive text corpora are now powerful tools to anticipate sequences of words (\emph{cf}.~GPT and variants~\cite{radford2018improving,radford2019language,brown2020language}). The incremental temporal modeling aspect has been also been explored for action recognition~\cite{li2020directional}, albeit with convolutional architectures and without  intermediate self-supervised losses.

While the architecture described so far can be applied on top of various frame or clip encoders (as we will show in experiments), we further propose a  \emph{purely attention-based video modeling architecture} by replacing the backbone with an attention-based frame encoder %
from the recently introduced Vision Transformer~\cite{dosovitskiy2021image}. This enables \method to attend not only to specific frames, but also to spatial features within the frames in one unified framework. As we see in~\cref{fig:teaser}, when trained on egocentric video, the model spontaneously learns to attend to spatial features corresponding to hands and objects, %
which tend to be especially important in anticipating future activities~\cite{liu2020forecasting}. %

In summary, our contributions are: 1) \method, a novel end-to-end purely attention based architecture  for predictive %
video modeling;
2) Incorporation of a self-supervised future prediction %
loss, making the architecture especially applicable to predictive %
tasks like action anticipation;
3) Extensive analysis and ablations of the model showing its versatility  with different backbone architectures, pre-trainings, \etc on the most popular action anticipation benchmarks, both from first and third person viewpoints. Specifically, we outperform all published prior work on \ekfull{}\footnote{\label{ekfootnote}\ekfull/100 datasets are licensed under the Creative Commons Attribution-NonCommercial 4.0 International License.}~\cite{Damen2018EPICKITCHENS}, 
\eknewfull{}\footref{ekfootnote}~\cite{damen2020rescaling}, \egtea~\cite{li2018eye}, and \saladfull~\cite{stein2013combining}. 
Most notably, %
our method outperforms all submissions to the \eknewfull CVPR'21 challenge\footnote{\href{https://competitions.codalab.org/competitions/25925}{\tt competitions.codalab.org/competitions/25925}}, and is ranked \#1 on the \ekfull leaderboard\footnote{\href{https://competitions.codalab.org/competitions/20071}{\tt competitions.codalab.org/competitions/20071}} for seen (S1) and \#2 on unseen (S2) test sets. %
\section{Related Work}

{\noindent \bf Action anticipation} is the task of predicting future actions given a video clip. %
While %
well explored in third-person video~\cite{stein2013combining,kuehne2014language,huang2014action,koppula2015anticipating,vondrick2016anticipating,jain2016recurrent,gao2017red,abu2018will}, %
it has recently gained in popularity for first-person (egocentric) videos~\cite{Damen2018EPICKITCHENS,sener2020temporal,liu2020forecasting,furnari2020rulstm,nagrajan2020egotopo,damen2020rescaling,dessalene2021forecasting}, due to its applicability on wearable computing platforms. Various approaches have been proposed for this task, such as learning representations by predicting future features~\cite{vondrick2016anticipating,wu2021imaginernn}, aggregating past features~\cite{furnari2020rulstm,sener2020temporal}, or leveraging affordances and hand motion~\cite{nagrajan2020egotopo,liu2020forecasting}. Our work contributes a new video architecture for anticipation, and we demonstrate its promising advantages on 
multiple
popular anticipation benchmarks.

{\noindent \bf Self-supervised feature learning from video} methods learn representations from unlabeled video, often to be fine-tuned for particular downstream tasks.  Researchers explore a variety of ``free'' supervisory signals, such as temporal consistency~\cite{jayaraman2016slow,fernando2017self,wei2018learning,kim2019self,yang2020vthcl}, inter-frame predictability~\cite{jayaraman2015learning,han2019dpc,sun2019contrastive,han2020memdpc}, and cross-modal correspondence~\cite{look-listen-2017,korbar2018cooperative,sun2019contrastive,sun2019videobert}.
\method incorporates losses that encourage features predictive of future features (and actions); while this aspect shares motivation with prior~\cite{vondrick2016anticipating,han2019dpc,han2020memdpc,sun2019contrastive,sun2019videobert,rodriguez2018action,liu2018future,luc2018predicting,shi2018action,gammulle2019predicting} and concurrent work~\cite{wu2021imaginernn}, our architecture to achieve predictive features is distinct (transformer based rather than convolutional/recurrent~\cite{han2019dpc,han2020memdpc,wu2021imaginernn,shi2018action,gammulle2019predicting}), it operates over raw frames or continuous video features as opposed to clustered `visual words'~\cite{sun2019videobert}, assumes only visual data (rather than vision with speech or text~\cite{sun2019contrastive,sun2019videobert}), and is jointly trained for action anticipation (rather than pre-trained and then fine-tuned for action recognition~\cite{han2019dpc,han2020memdpc,sun2019contrastive}).

{\noindent \bf Language modeling (LM)} has been revolutionized with the introduction of self-attention architectures~\cite{vaswani2017attention}. LM approaches can generally be classified in three categories: (1) {\em encoder-only}~\cite{devlin2019bert,peters2018deep}, which leverage bidirectional attention and are effective for discriminative tasks such as classification; (2) {\em decoder-only}~\cite{brown2020language,radford2018improving}, which leverage a causal attention~\cite{lample2019cross} attending on past tokens, and are effective for generative tasks such as text generation; and (3) {\em encoder-decoder}~\cite{raffel2020exploring,lewis2020bart}, which incorporate both a bidirectional encoder and causal decoder, and are effective for tasks such as machine translation. %
Capitalizing on the
analogy between
action prediction and generative language tasks, we explore causal decoder-only attention architectures in our model.  While language models are typically trained on discrete inputs (words), \method trains with continuous video features.  This distinction naturally influences our design choices, such as an $L_2$ loss for generative training as opposed to a cross entropy loss for the next word.
{\noindent \bf Self-attention and transformers in vision.}
The general idea of self-attention in vision dates back to non-local means~\cite{buades2005non}, and is incorporated into contemporary network architectures as non-local blocks~\cite{wang2017non,wu2019long,liu2019non,cao2019gcnet} and gating mechanisms~\cite{kong2017low,xie2018rethinking,girdhar2017attentional,miech17loupe}. While self-attention approaches like transformers~\cite{vaswani2017attention,wang2019learning} offer %
strong results for high-level vision reasoning tasks~\cite{carion2020end,zhao2020point}, more recently, there %
is growing interest in completely replacing convolutional architectures with transformers for image recognition~\cite{dosovitskiy2021image,touvron2020deit}. %
For video, prior work 
has mostly leveraged attention architectures~\cite{wu2019long,wang2017non,girdhar2019video} on top of 
standard spatiotemporal convolutional base architectures~\cite{carreira2017quo,tran2015learning,tran2018closer}.
In contrast, \method is an end-to-end transformer architecture for video---to our knowledge the first (concurrent with~\cite{bertasius2021is,neimark2021video,arnab2021vivit,fan2021multiscale,li2021vidtr}).
Unlike the concurrent methods~\cite{bertasius2021is,neimark2021video,arnab2021vivit,fan2021multiscale,li2021vidtr}, which 
are bidirectional and address traditional action recognition,
\method has %
a causal structure and tackles predictive tasks (anticipation). %
\method yields 
the best results to date for several well-studied anticipation benchmarks.  

\begin{figure}[t]
    \centering
    \includegraphics[width=\linewidth,bb=0 0 525 72]{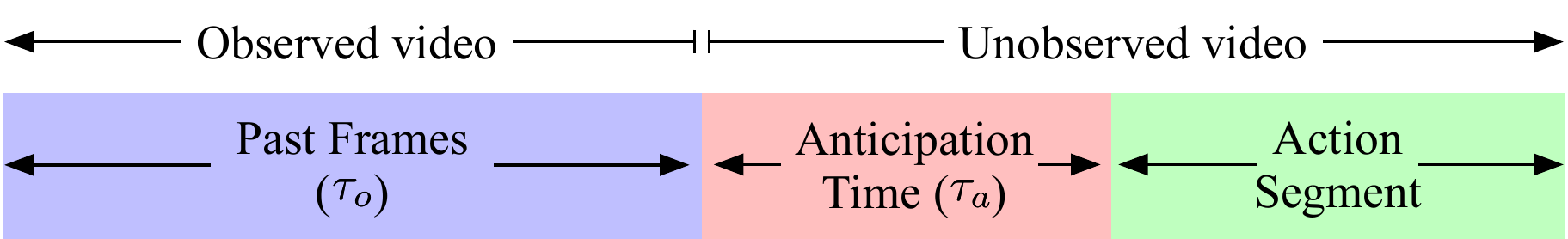}
    \caption{{\bf Action anticipation problem setup.} The goal is to use the observed video segment of length $\tau_o$ to anticipate the future action $\tau_a$ seconds before it happens.}\label{fig:problem_setup}
\end{figure}

\section{Anticipation Problem Setup}\label{sec:app:prob_setup}

While multiple anticipation problem setups have been explored in the literature~\cite{kitani2012activity,rhinehart2017first,nagrajan2020egotopo}, in this work we follow the setup defined in recent challenge benchmarks~\cite{Damen2018EPICKITCHENS,damen2020rescaling} and illustrated in~\cref{fig:problem_setup}. For each action segment labeled in the dataset starting at time $\tau_s$, the goal is to recognize it using a $\tau_o$ length video segment $\tau_a$ units before it, \ie from $\tau_s - (\tau_a + \tau_o)$ to $\tau_s - \tau_a$. While methods are typically allowed to use any length of observed segments ($\tau_o$), the anticipation time ($\tau_a$) is usually fixed for each dataset. 

\begin{figure*}[t]
    \centering
    \raisebox{-0.5\height}{\includegraphics[width=0.84\linewidth,bb=7 8 885 504]{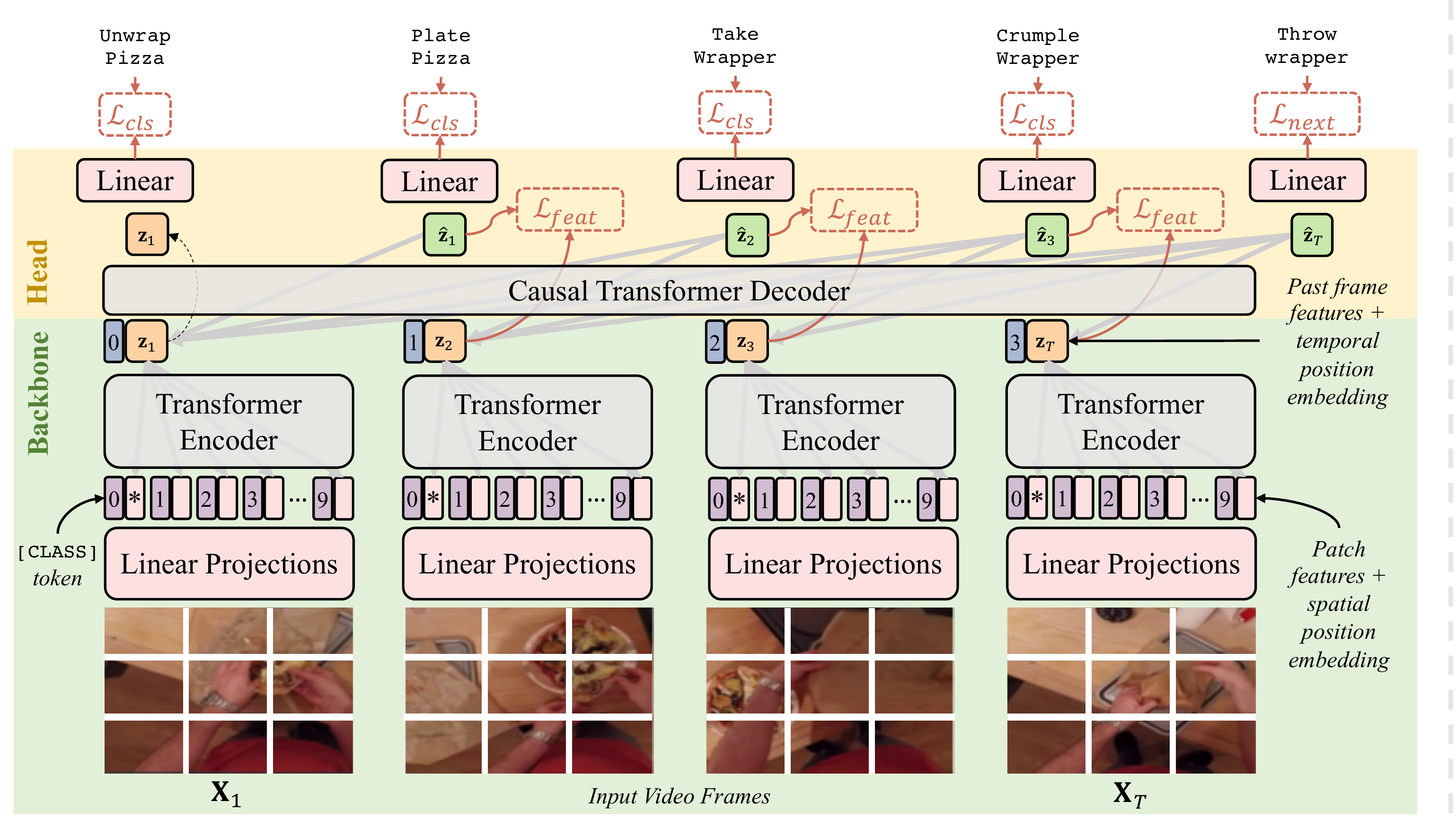}} \hfill
    \raisebox{-0.5\height}{\includegraphics[width=0.14\linewidth,bb=6 0 194 653]{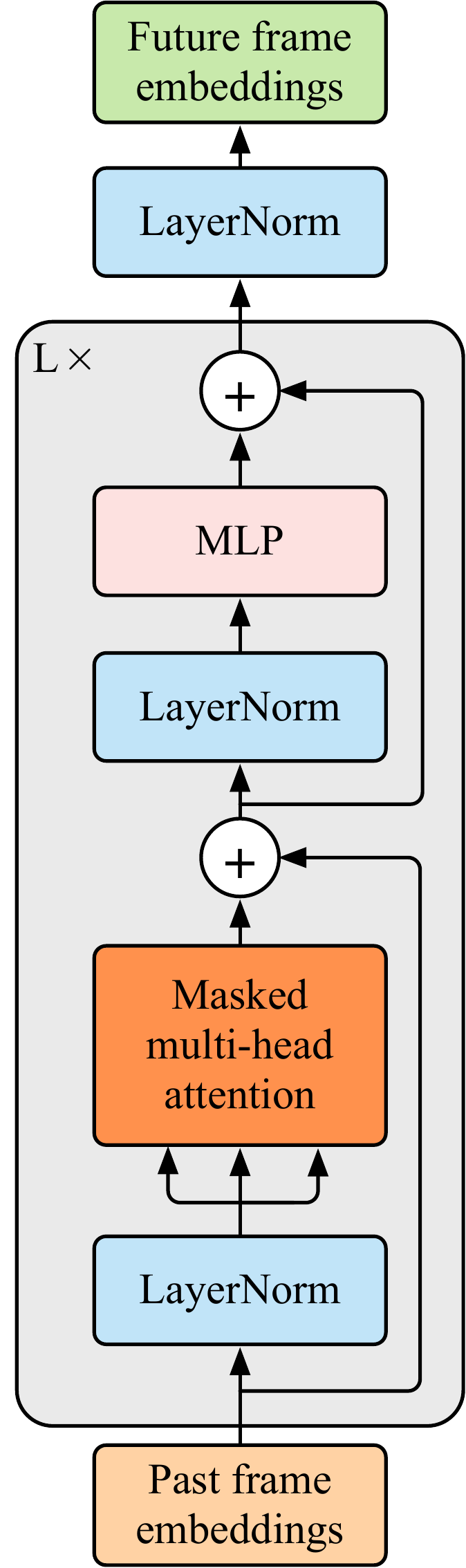}}
    \caption{
        {\em (Left)} {\bf \method architecture.} We split the $T$ input frames into non-overlapping patches that are linearly projected. We add %
        a learned \texttt{[CLASS]} token, along with spatial position embeddings, and the resulting features are passed through multiple layers of multi-head attention, with shared weights across the transformers applied to all frames. We take the resulting features corresponding to the \texttt{[CLASS]} token, append a temporal position encoding  and pass it through the Causal Transformer Decoder that predicts the future feature at frame $t$, after attending to all features from $1\cdots t$. The resulting feature is trained to regress to the true future feature ($\lossGPT$) and predict the action at that time point if labeled ($\lossPast$), and the last prediction is trained to predict the future action ($\lossFuture$). {\em (Right)} {\bf Causal Transformer Decoder.} It follows the Transformer architecture with pre-norm~\cite{wang2019learning}, causal masking in attention, and a final LayerNorm~\cite{radford2019language}. 
        }
        \label{fig:approach}
\end{figure*}

\section{\methodfull{}}

We now present the \method model architecture, as illustrated in~\cref{fig:approach}. It is designed to predict future actions given a video clip as input. To that end, it leverages a two-stage architecture, consisting of a {\em backbone} network that operates on individual frames or short clips, followed by a {\em head} architecture that operates on the frame/clip level features to predict future features and actions. \method employs {\em causal} attention modeling---predicting the future actions based only on the frames observed so far---and is trained using objectives inspired from self-supervised learning. We now 
describe
each model component in detail, followed by the training and implementation details.

\subsection{Backbone Network}

Given a video clip with $T$ frames, $V=\{{\bf X}_1, \cdots, {\bf X}_T\}$
the backbone network, $\calB$, extracts a feature representation for each frame, $\{{\bf z}_1, \cdots, {\bf z}_T\}$ where ${\bf z}_t = \calB({\bf X}_t)$.
While various video base architectures have been proposed~\cite{carreira2017quo,tran2019video,feichtenhofer2019slowfast,wang2016tsn} and can be used with \method as we demonstrate later, in this work we propose an alternate architecture for video understanding based purely on attention. This backbone, which we refer to as \txBack, adopts the recently proposed Vision Transformer (ViT)~\cite{dosovitskiy2021image} architecture, which has shown impressive results for static image classification. 

Specifically, we adopt the ViT-B/16 architecture. We split each input frame into $16\times 16$ non-overlapping patches. We flatten each patch into a 256D vector, and linearly project them to 768D, which is the feature dimension used throughout the encoder. While we do not need to classify each frame individually, we still prepend a learnable {\tt [class]} token embedding to the patch features, 
whose output will be used as a frame-level embedding %
input to the head. Finally, we add learned position embeddings 
to each patch feature similar to~\cite{dosovitskiy2021image}. %
We choose to stick to frame-specific spatial position encodings, so that the same backbone model with shared weights can be applied to each frame. We will incorporate the temporal position information in the head architecture (discussed next).
The resulting patch embeddings are passed through a standard Transformer Encoder~\cite{vaswani2017attention} with pre-norm~\cite{wang2019learning}. 
We refer the reader to~\cite{dosovitskiy2021image} for details of the encoder architecture.

\txBack is an attractive backbone design because it makes our architecture purely attentional.  Nonetheless, in addition to \txBack, \method is compatible with other video backbones, including those based on 2D CNNs~\cite{wang2016tsn,Simonyan_14b}, 3D CNNs~\cite{carreira2017quo,tran2019video,feichtenhofer2019slowfast}, or fixed feature representations based on detected objects~\cite{bertasius2020classifying,bertasius2020cobe} or visual attributes~\cite{miech2019leveraging}. In~\cref{sec:expt} we provide experiments testing several such alternatives.  For the case of spatiotemporal backbones, which operate on clips as opposed to frames, we extract features as ${\bf z}_t = \calB({\bf X}_{t-L}, \cdots, {\bf X}_t)$, where the model is trained on $L$-length clips.  This ensures the features at frame $t$ do not incorporate any information from the future, which is not allowed in the anticipation problem setting.  %

\subsection{Head Network}

Given the features extracted by the backbone, the head network, referred to as \txHead, is used to predict the future features for each input frame using a Causal Transformer Decoder, $\calD$: 
\begin{equation}
    {\bf \hat{z}}_{1}, \cdots, {\bf \hat{z}}_{T} = \calD({\bf z}_1, \cdots, {\bf z}_T).
    \end{equation}
Here ${\bf \hat{z}}_{t}$ is the predicted future feature corresponding to frame feature ${\bf z}_{t}$, after attending to all features before and including it.
The predicted features are then decoded into a distribution over the semantic action classes using a linear classifier $\theta$, \ie ${\bf \hat{y}}_t = \theta({\bf \hat{z}}_t)$. The final prediction, ${\bf \hat{y}}_{T}$, is used as the model's output for the next-action anticipation task.  
Note that since the next action segment ($T+1$) is $\tau_a$ seconds from the last observed frame ($T$) as per the problem setup, we typically sample frames at a stride of $\tau_a$ so that the model learns to predict future features/actions at that frame rate. However, empirically we find the model is robust to other frame rate values as well.

We implement $\calD$  using a masked transformer decoder inspired from popular approaches in generative language modeling, such as GPT-2~\cite{radford2019language}.
We start by adding a temporal position encoding to the frame features implemented as a learned embedding of the absolute frame position within the clip. The embedded features are then passed through multiple decoder layers, 
each consisting of masked multi-head attention, LayerNorm (LN) and a multi-layer perceptron (MLP), as shown in~\cref{fig:approach} (right).
The final output is then passed through another LN, akin to GPT-2~\cite{radford2019language}, to obtain the future frame embeddings.  

Aside from being visual rather than textual, this model differs from the original Transformer Decoder~\cite{vaswani2017attention} in terms of the final LN and the masking operation in the multi-head attention. The masking ensures that the model only attends to specific parts of the input, which in the case of
predictive tasks like ours, is defined as a `causal' mask. That is, for the output corresponding to the future after frame $t$, \ie ${\bf \hat{z}}_t$, we set the mask to only attend to ${\bf z}_1 \cdots {\bf z}_t$. 
We refer the reader to~\cite{radford2019language} for details on the masking implementation.
This design differs %
considerably from previous applications of language modeling architectures to video, such as VideoBERT~\cite{sun2019videobert}. It operates directly on continuous clip embeddings instead of first clustering them into tokens, and it leverages causal attention to allow for \lossBoth training (discussed next), instead of needing masked language modeling (MLM) as in BERT~\cite{devlin2019bert}. 
These properties make \method suited for predictive video tasks while allowing for the long-range reasoning %
that is often lost in recurrent architectures.
While follow-ups to VideoBERT such as CBT~\cite{sun2019contrastive} operate on raw clip features, they still leverage a MLM objective with bidirectional attention, with the primary goal of representation learning as opposed to future prediction.

\subsection{Training \method} %

To sample training data, for each labeled action segment in a given dataset, we sample a clip preceding it and ending $\tau_a$ seconds before the start of the action. %
We pass the clip through \method to obtain future predictions, and then supervise the network using three losses. 

First, we supervise the next-action prediction using a cross-entropy loss with the labeled future action, $c_{T+1}$:
\begin{align}
    \lossFuture = -\log {\bf \hat{y}}_{T}[c_{T+1}].
\end{align}
Second, to leverage the causal structure of the model, we supervise the model's intermediate future predictions at the \emph{feature} level and the \emph{action class} level.  %
For the former, we predict future features to match the true future features that are present in the clip, \ie
\begin{align}
    \lossGPT = \sum_{t=1}^{T-1} || {\bf \hat{z}}_{t} - {\bf z}_{t+1} ||_{2}^{2}.
\end{align}
This loss is inspired from the seminal work by Vondrick \etal~\cite{vondrick2016anticipating} as well as follow ups~\cite{han2019dpc,han2020memdpc} that show that anticipating future visual representations %
is an effective form of self-supervision, though typically for traditional action recognition tasks.
Concurrent and recent work adopts similar objectives for anticipation tasks, but with recurrent architectures~\cite{wu2021imaginernn,shi2018action,gammulle2019predicting}.  Whereas recent methods~\cite{han2019dpc,han2020memdpc,wu2021imaginernn} explore this loss with NCE-style~\cite{oord2018representation} objectives, in initial experiments we found simple $L_2$ loss to be equally effective. Since our models are always trained with the final supervised loss, we do not suffer from potential collapse during training that would necessitate the use of contrastive losses.

Third, as an \emph{action class} level \lossBoth loss, we leverage any action labels available in the dataset to supervise the intermediate predictions, \ie, when the input clip overlaps with any labeled action segments that precede the segment to be anticipated.\footnote{For example, this would be true for each frame for densely labeled datasets like \saladfull, and a subset of frames for sparsely labeled datasets like \ekfull.} Setting $c_t=-1$ for any earlier frames for which we do not have labels, we incur the following loss:
\begin{align}
    \lossPast &= \sum_{t=1}^{T-1} \lossPast^t ; \quad
    \lossPast^{t} &=
    \begin{cases}
        -\log {\bf \hat{y}}_{t}[c_{t+1}] & \text{if } c_{t+1}\geq 0\\
        0              & \text{otherwise}.
    \end{cases}
\end{align}
We train our model with 
\begin{equation}
    \calL = \lossFuture + \lossPast + \lossGPT
\end{equation}
as the objective, and refer to it as the {\bf \causalSetting} \causalSettingShort training setting. %
As a baseline, we also experiment with a model trained solely with $\calL = \lossFuture$, and refer to it as the {\bf \acausalSetting} \acausalSettingShort setting, as it does not leverage our model's causal attention structure, instead supervising only the final prediction which attends to the full input. 
As we will show in~\cref{tab:abl:losses}, the \causalSetting setting leads to significant improvements. %

\subsection{Implementation Details}

We preprocess the input video clips by randomly scaling the height between 248 and 280px, and take 224px crops at training time. We sample 10 frames at 1FPS for most experiments.
We adopt network architecture details from~\cite{dosovitskiy2021image} for the \txBack backbone. Specifically, we use a 12-head, 12-layer transformer encoder model that operates on 768D representations. We initialize the weights from a model pre-trained on ImageNet-1K (\imnet), ImageNet-21K (\imnetii) or ImageNet-1K finetuned from ImageNet-21K (\imnetboth), and finetune end-to-end for the anticipation tasks. For \txHead, we use a 4-head, 6-layer model that operates on a 2048D representation, initialized from scratch. We employ a linear layer between the backbone and head to project the features to match the feature dimensions used in the head. We train \method end-to-end with SGD+momentum using $10^{-6}$ weight decay and $10^{-4}$ learning rate for 50 epochs, with a 20 epoch warmup~\cite{goyal2017accurate} and 30 epochs of cosine annealed decay. 
At test time, we employ 3-crop testing, where we compute three 224px  spatial crops from 248px input frames, and average the predictions over the corresponding three clips.
The default backbone for \method is \txBack, based on the ViT-B/16 architecture.  However, to enrich our comparisons with some baselines~\cite{furnari2020rulstm,furnari2019rulstm,sener2020temporal}, below also we report performance of only our head model operating on fixed features from  1) a frame-level TSN~\cite{wang2016tsn} backbone pre-trained for action classification, or 2) a recent spatiotemporal convolutional architecture irCSN-152~\cite{tran2019video} pre-trained on a large weakly labeled video dataset~\cite{ghadiyaram2019large}, which has shown strong results when finetuned for action recognition. We finetune that model for action classification on the anticipation dataset 
and extract features that are used by the head for anticipation. 
In these cases, we only train the \txHead layers.
For all datasets considered, we use the validation set or split 1 to further optimize the hyperparameters, and use that setup over multiple splits or the held out test sets.
Code and models will be released for reproducibility.
\section{Experiments}\label{sec:expt}

We empirically evaluate \method on four popular action anticipation benchmarks covering both first- and third-person videos. We start by describing the datasets and evaluation protocols (\cref{sec:expt:setup}), followed by key results and comparisons to the state of the art (\cref{sec:expt:sota}), and finally ablations and qualitative results (\cref{sec:expt:ablations}).

\subsection{Experimental Setup}\label{sec:expt:setup}

\begin{table}[t]
\centering
\setlength{\tabcolsep}{3pt}
\resizebox{\columnwidth}{!}{%
\begin{tabular}{@{}llllll@{}}
\toprule
Dataset & Viewpoint & Segments & Classes & $\tau_a$ (s) & Metric(s) \\ 
\midrule
\eknew~\cite{damen2020rescaling} & 1st & 90.0K & 3,807 & 1.0~\cite{damen2020rescaling} & recall \\
\ek~\cite{Damen2018EPICKITCHENS} & 1st & 39.6K & 2,513 & 1.0~\cite{Damen2018EPICKITCHENS} & top-1/5, recall\\
\egtea~\cite{li2018eye} & 1st & 10.3K & 106 & 0.5~\cite{liu2020forecasting} & top-1, cm top-1 \\
\salad~\cite{stein2013combining} & 3rd & 0.9K & 17  & 1.0~\cite{abu2018will} & top-1\\
\bottomrule
\end{tabular}}
\caption{{\bf Datasets} used for evaluation. We use four popular benchmarks, spanning first and third person videos. Class-mean (`cm') $\implies$evaluation is done per-class and averaged over classes. Recall refers to class-mean recall@5 from~\cite{furnari2018leveraging}. For all, higher is better.
}
\label{tab:datasets}
\end{table}

{\noindent \bf Datasets and metrics.}
We test on four popular action anticipation datasets summarized in~\cref{tab:datasets}.
{\em \eknewfull (\eknew)}~\cite{damen2020rescaling} is the largest egocentric (first-person) video dataset with 700 long unscripted videos of cooking activities totalling 100 hours. 
{\em \ekfull (\ek)}~\cite{Damen2018EPICKITCHENS} is an earlier version of the same, and allows for comparisons to a larger set of baselines which have not yet been reported on \eknew.
For both, we use the standard train, val, and test splits from~\cite{damen2020rescaling} and~\cite{furnari2019rulstm} respectively to report performance. The test evaluation is performed on a held-out set through a submission to their challenge server.
{\em \egtea}~\cite{li2018eye} is another popular egocentric action anticipation dataset.
Following recent work~\cite{liu2020forecasting}, we report performance on the split 1~\cite{li2018eye} of the dataset at $\tau_a=0.5s$. %
Finally, %
{\em \saladfull (\salad)}~\cite{stein2013combining} is a popular third-person anticipation dataset, and we report top-1 accuracy
averaged over the pre-defined 5 splits %
following prior work~\cite{sener2020temporal}.
Some of these datasets employ top-5/recall@5 criterion to account for the multi-modality in future predictions, as well as class-mean (cm) metrics to equally weight classes in a long-tail distribution.
The first three datasets also decompose the action annotations into verb and nouns.
While some prior work~\cite{sener2020temporal} supervises the model additionally for nouns and verbs,
we train all our model solely to predict actions, and estimate the verb/noun probabilities by marginalizing over the other, similar to~\cite{furnari2019rulstm}.
In all tables, we highlight the columns %
showing the metric used to rank methods in the official challenge leaderboards. 
Unless otherwise specified, the reported metrics correspond to future action (act.) prediction, although we do report numbers for verb and nouns separately where applicable.
Please see~\cref{sec:appdx:dataset} for further details.

{\noindent \bf Baselines.} 
We compare \method to its variants with different backbones and pretrained initializations, as well as to the strongest recent approaches for action anticipation,
\ie \rulstm~\cite{furnari2019rulstm,furnari2020rulstm}, \acbank~\cite{sener2020temporal}, and Forecasting HOI (\fhoi)~\cite{liu2020forecasting}. 
Please see~\cref{sec:appdx:baselines} for details on them.
While \fhoi trains the model end-to-end, \rulstm and \acbank operate on top of features from a model pretrained for action classification on that dataset. Hence, we
report results both using the exact same features as well as end-to-end trained  backbones  
to facilitate fair comparisons.  

\subsection{Comparison to the \sota}\label{sec:expt:sota}

\begin{table}[t]
\centering
\setlength{\tabcolsep}{3pt}
\resizebox{\columnwidth}{!}{%
\begin{tabular}{HlHllccg}
\toprule
&Head & Sampling & Backbone & Init & Verb & Noun & Action \\
\midrule
\multirow{5}{*}{\rotatebox{90}{RGB}} 
& \rulstm~\cite{damen2020rescaling} %
& \rulstm & TSN %
& \imnet & 27.5 & 29.0 & 13.3 \\
&\txHead & \correctSample & TSN %
& \imnet & 27.2 & 30.7 & 13.6 \\ %
& \txHead & \correctSample & irCSN152 %
& IG65M & 25.5 & 28.1 & 12.8 \\ %
& \txHead & \correctSample & \txBack{} & \imnet & 28.2 & 29.3 & 13.4 \\ %
& \txHead & \correctSample & \txBack{} & \imnetboth & 28.7 & {\bf 32.3} & 14.4 \\ %
& \txHead & \correctSample & \txBack{} & \imnetii & {\bf 30.2} & 31.7 & {\bf 14.9} \\ %
\arrayrulecolor{gray}
\midrule
\multirow{2}{*}{\rotatebox{90}{OBJ}}
& \rulstm~\cite{damen2020rescaling} & \rulstm & Faster R-CNN %
& \imnet & 17.9 & 23.3 & 7.8 \\
& \txHead & \correctSample & Faster R-CNN %
& \imnet & {\bf 18.0} & {\bf 24.3} & {\bf 8.7} \\  %
\arrayrulecolor{black}
\bottomrule
\end{tabular}%
}
\caption{{\bf \eknew (val)} using RGB and detected objects (OBJ) modalities separately. \method outperforms prior work using the exact same features, and further improves with our \txBack backbone. 
Performance reported using class-mean recall@5. %
}
\label{tab:epic100_rgb}
\end{table}

\begin{table}[t]
\centering
\setlength{\tabcolsep}{3pt}
\resizebox{\columnwidth}{!}{%
\begin{tabular}{clccgcccccc}
\toprule
& & \multicolumn{3}{c}{Overall} & \multicolumn{3}{c}{Unseen Kitchen} & \multicolumn{3}{c}{Tail Classes} \\
\cmidrule(lr){3-5} \cmidrule(lr){6-8} \cmidrule(lr){9-11}
Split & Method & \multicolumn{1}{c}{Verb} & \multicolumn{1}{c}{Noun} & \multicolumn{1}{c}{Act} & \multicolumn{1}{c}{Verb} & \multicolumn{1}{c}{Noun} & \multicolumn{1}{c}{Act} & \multicolumn{1}{c}{Verb} & \multicolumn{1}{c}{Noun} & \multicolumn{1}{c}{Act} \\ \midrule
\multirow{4}{*}{\rotatebox{90}{Val}}
&chance& 6.4 & 2.0 & 0.2 & 14.4 & 2.9 & 0.5 & 1.6 & 0.2 & 0.1 \\
&\rulstm~\cite{damen2020rescaling}
& 27.8 & 30.8 & 14.0 & 28.8 & {\bf 27.2} & {\bf 14.2} & 19.8 & 22.0 & 11.1\\
& \methodfused (TSN) & 25.5 & 31.8 & 14.8 & 25.5 & 23.6 & 11.5 & 18.5 & 25.8 & 12.6 \\  %
& \methodfused &  {\bf 28.2} & {\bf 32.0} & {\bf 15.9} & {\bf 29.5} & 23.9 & 11.9 & {\bf 21.1} & {\bf 25.8} & {\bf 14.1} \\  %
\arrayrulecolor{black}
\cmidrule{1-11} \multirow{3}{*}{\rotatebox{90}{Test}}
&chance & 6.2 & 2.3 & 0.1 & 8.1 & 3.3 & 0.3 & 1.9 & 0.7 & 0.0 %
\\
&\rulstm{}~\cite{damen2020rescaling}& 25.3 & 26.7 & 11.2 & 19.4 & 26.9 & 9.7 & 17.6 & 16.0 & 7.9 \\
& TBN~\cite{zatsarynna2021multimodal} & 21.5 & 26.8 & 11.0 & 20.8 & {\bf 28.3} & {\bf 12.2} & 13.2 & 15.4 & 7.2 \\
& \methodfused & {\bf 25.6} & {\bf 28.8} & {\bf 12.6} & {\bf 20.9} & 22.3 & 8.8 & {\bf 19.0} & {\bf 22.0} & {\bf 10.1} \\ %
\arrayrulecolor{gray}
\cmidrule{2-11} \multirow{5}{*}{\rotatebox{90}{Challenge}}
& IIE\_MRG & 25.3 & 26.7 & 11.2 & 19.4 & 26.9 & 9.7 & 17.6 & 16.0 & 7.9 \\
& NUS\_CVML~\cite{sener2021technical} & 21.8 & 30.6 & 12.6 & 17.9 & 27.0 & 10.5 & 13.6 & 20.6 & 8.9 \\
& ICL+SJTU~\cite{gu2021epicantchallenge} & {\bf 36.2} & 32.2 & 13.4 & {\bf 27.6} & 24.2 & 10.1 & {\bf 32.1} & {\bf 29.9} & 11.9 \\
& Panasonic~\cite{yamamuro2021epicantchallenge} & 30.4 & {\bf 33.5} & 14.8 & 21.1 & 27.1 & 10.2 & 24.6 & 27.5 & 12.7 \\
& \methodfusedall & 26.7 & 32.3 & {\bf 16.7} & 21.0 & {\bf 27.6} & {\bf 12.9} & 19.3 & 24.0 & {\bf 13.8} \\ %
\arrayrulecolor{black}
\bottomrule
\end{tabular}}
\caption{{\bf \eknew val and test sets} using all modalities. We split the test comparisons between published work and CVPR'21 challenge submissions. We outperform prior work including all challenge submissions, with especially significant gains on tail classes.
Performance is reported using class-mean recall@5.
\methodfused and \methodfusedall late fuse predictions from multiple modalities; please see text for details.
}
\label{tab:ek100}
\end{table}

{\noindent \bf \eknew.}
We first compare \method to prior work using individual modalities (RGB and Obj~\cite{furnari2019rulstm}) in \cref{tab:epic100_rgb} for apples-to-apples comparisons and to isolate the performance of each of our contributions. 
First, we compare to the \sota \rulstm method using only our \method (head) model applied to the exact same features from TSN~\cite{wang2016tsn} trained for classification on \eknew. We note this already improves over \rulstm, particularly in anticipating future objects (nouns). 
Furthermore, we experiment with backbone features from a recent \sota video model, irCSN-152~\cite{tran2019video} pretrained on a large weakly supervised dataset, IG65M~\cite{ghadiyaram2019large}. %
We finetune this backbone for recognition on \eknew, extract its features and train \txHead same as before, but find it to not be particularly effective at the \eknew anticipation task.
Next, we replace the backbone with our \txBack and train the model end-to-end, leading to the best performance so far, and outperforming \rulstm by 1.6\%. We make the same comparison over features from an object-detector~\cite{ren2015faster} trained on \eknew provided by \rulstm (referred to as OBJ modality, details in~\cref{sec:appdx:dataset}), and similarly find our method outperforms \rulstm on this modality as well. 

Note that the fixed features used above can be thought of as a proxy for past recognized actions, as they are trained only for action recognition. Hence, \txHead on TSN or irCSN152 features is comparable to a baseline that trains a language model over past actions to predict future ones. As the later experiments show, end-to-end trained \method is significantly more effective, supporting \method's \emph{from-pixels} anticipation as opposed to \emph{label-space} anticipation. %
Finally, we compare models using all modalities on the \eknew val and the held-out test set in~\cref{tab:ek100}. While \rulstm fuses models trained on RGB, Flow, and OBJ features using an attention based model (MATT~\cite{furnari2019rulstm}), we simply late fuse predictions from our best RGB and OBJ models (resulting model referred to as \methodfused), and outperform all reported work on this benchmark, establishing a new \sota. Note we get the largest gains on tail classes, suggesting our model is particularly effective at few-shot anticipation. 
Finally, \methodfusedall ensembles multiple model variants, and outperforms all submissions on the \eknew CVPR'21 challenge leaderboard. Please refer to the workshop paper~\cite{girdhar2021avtchallenge} for details on \methodfusedall.

\begin{table}[t]
\centering
\setlength{\tabcolsep}{3pt}
\resizebox{\columnwidth}{!}{%
\begin{tabular}{lHll|gc|c}
\toprule
Head & Sampling & Backbone & Init &  Top-1 & Top-5 & Recall \\
\midrule
\rulstm~\cite{furnari2020rulstm} & \rulstm & TSN & \imnet & 13.1 & 30.8 & 12.5 \\
\acbank~\cite{sener2020temporal} & \acbank & TSN & \imnet & 12.3 & 28.5 & 13.1 \\ %
\arrayrulecolor{gray}
\midrule
\txHead  & \correctSample & TSN & \imnet & 13.1 & 28.1 & 13.5 \\  %
\txHead & \correctSample & \txBack{} & \imnetboth & 12.5 & 30.1 & {\bf 13.6} \\ %
\txHead & \correctSample & irCSN152 & IG65M & {\bf 14.4} & {\bf 31.7} & 13.2 \\ %
\arrayrulecolor{black}
\bottomrule
\end{tabular}%
}
\caption{{\bf \ek using only RGB modality} for action anticipation. \method performs comparably, and outperforms when combined with a backbone pretrained on  large weakly labeled dataset.
}
\label{tab:epic55_rgb}
\end{table}

{\noindent \bf \ek.}
Since \eknew is relatively new and has few baseline methods reported, we also evaluate \method on \ek. As before, we start by comparing single modality methods (RGB-only) in~\cref{tab:epic55_rgb}. 
For \txHead models, we found a slightly different set of (properly validated) hyperparameters performed better for top-1/5 metrics vs.\ the recall metric, hence we report our best models for each set of results.
Here we find \txHead performs comparably to \rulstm, and outperforms another attention-based model~\cite{sener2020temporal} (one of the winners of the \ek 2020 challenge) on the top-1 metrics. 
The gain is more significant on the recall metric, which averages performance over classes, indicating again that \txHead is especially effective on tail classes which get ignored in top-1/5 metrics.
Next, we replace the backbone with \txBack, and find it to perform comparably %
on top-1/5 metrics, and outperforms on the recall metric.
Finally, we experiment with irCSN-152~\cite{tran2019video} pretrained using IG65M~\cite{ghadiyaram2019large} and finetuned on \ek, and find it to outperform all methods by a significant margin on top-1/5. We show further comparisons with the \sota on \ek in~\cref{sec:appdx:ek55_full}.
\begin{table}[t]
    \centering
    \setlength{\tabcolsep}{2pt}
    \begin{minipage}[t]{0.65\linewidth}
    \resizebox{\columnwidth}{!}{%
    \begin{tabular}[t]{lccgccg}
    \toprule
    & \multicolumn{3}{c}{Top-1 acc.} & \multicolumn{3}{c}{Class mean acc.} \\
    \cmidrule(r){2-4} \cmidrule(lr){5-7}
    Method & \multicolumn{1}{c}{Verb} & \multicolumn{1}{c}{Noun} & \multicolumn{1}{c}{Act.} & \multicolumn{1}{c}{Verb} & \multicolumn{1}{c}{Noun} & \multicolumn{1}{c}{Act.} \\ \midrule
    I3D-Res50~\cite{carreira2017quo} & 48.0 & 42.1 & 34.8 & 31.3 & 30.0 & 23.2 \\
    \fhoi~\cite{liu2020forecasting} & 49.0 & 45.5 & 36.6 & 32.5 & 32.7 & 25.3\\
    \arrayrulecolor{gray}
    \midrule
    \txHead (+TSN) & 51.7 & 50.3 & 39.8 & 41.2 & 41.4 & 28.3  \\ %
    \method & {\bf 54.9} & {\bf 52.2} & {\bf 43.0} & {\bf 49.9} & {\bf 48.3} & {\bf 35.2} \\ %
    \arrayrulecolor{black}
    \bottomrule
    \end{tabular}%
    }
    \caption{{\bf \egtea} Split 1 at $\tau_a=0.5s$. \method outperforms prior work by significant margins, especially when trained end-to-end with the \txBack backbone.
    }
    \label{tab:egtea_05}
    \end{minipage} \hfill
    \begin{minipage}[t]{0.33\linewidth}
    \resizebox{\columnwidth}{!}{%
    \begin{tabular}[t]{lHc}
    \toprule
    Head & Backbone/Input & Top-1 \\ 
    \midrule
    DMR~\cite{vondrick2016anticipating} & FC7 features & 6.2 \\
    RNN~\cite{abu2018will} & FV+segmentation~\cite{richard2017weakly} & 30.1 \\
    CNN~\cite{abu2018will} & FV+segmentation~\cite{richard2017weakly} & 29.8 \\
    \acbank~\cite{sener2020temporal} & I3D & 40.7 \\
    \arrayrulecolor{gray}
    \midrule
    \method %
    & \txBack{} & {\bf 48.0} \\ %
    \arrayrulecolor{black}
    \bottomrule
    \end{tabular}%
    }
    \caption{{\bf \saladfull.} \method{} outperforms prior work
    even in 3rd person videos. 
    }
    \label{tab:salad}
    \end{minipage}
\end{table}

{\noindent \bf \egtea.} In~\cref{tab:egtea_05} we compare our method at $\tau_a=0.5s$ on the split 1 as in recent work~\cite{liu2020forecasting}. Even using fixed features with \txHead on top, \method outperforms the best reported results, and using the \txBack backbone further improves performance. Notably, \fhoi leverages attention on hand trajectories to obtain strong performance, which, as we see in~\cref{fig:teaser}, emerges spontaneously in our model.

{\noindent \bf \saladfull.} Finally, we show that our approach is not limited to egocentric videos and is also effective in third-person settings. In~\cref{tab:salad}, we report top-1 performance on \saladfull averaged over standard 5 splits. We observe it outperforms previous RNN~\cite{abu2018will} and attention~\cite{sener2020temporal} based approaches by a significant 7.3\% absolute improvement, again establishing a new \sota.

\subsection{Ablations and Analysis}\label{sec:expt:ablations}

We now analyze the \method architecture, using the RGB modality and \eknew validation set as the test bed.

\begin{figure}[t]
\begin{minipage}[t]{0.52\linewidth}
    \vspace{0pt}
    \centering
    \setlength{\tabcolsep}{3pt}
    \resizebox{\linewidth}{!}{%
    \begin{tabular}{@{}lcccc@{}}
        \toprule
        & \multicolumn{2}{c}{Losses} & \multicolumn{2}{c}{Backbones} \\
        \cmidrule(r){2-3} \cmidrule(l){4-5}
        Setting & $\lossPast$ & $\lossGPT$ & TSN & \txBack{} \\ 
        \midrule
        \acausalSetting \acausalSettingShort & - & - & 10.1 & 13.1 \\
        & \cmark  & - & 11.5 & 14.4 \\
        & - & \cmark & 13.7 & 13.0 \\
        \causalSetting \causalSettingShort & \cmark  & \cmark & 13.6 & 14.4 \\
        \bottomrule
    \end{tabular}%
    }
    \vspace{0.015in}
    \captionof{table}{{\bf \LossBoth training.} Employing the \lossBoth training losses are imperative to obtain strong performance with \method. Reported on \eknew/cm recall@5.
    }\label{tab:abl:losses}
\end{minipage}\hfill
\begin{minipage}[t]{0.45\linewidth}
    \vspace{0pt}
    \centering
    \includegraphics[width=\linewidth,bb=0 1 627 447]{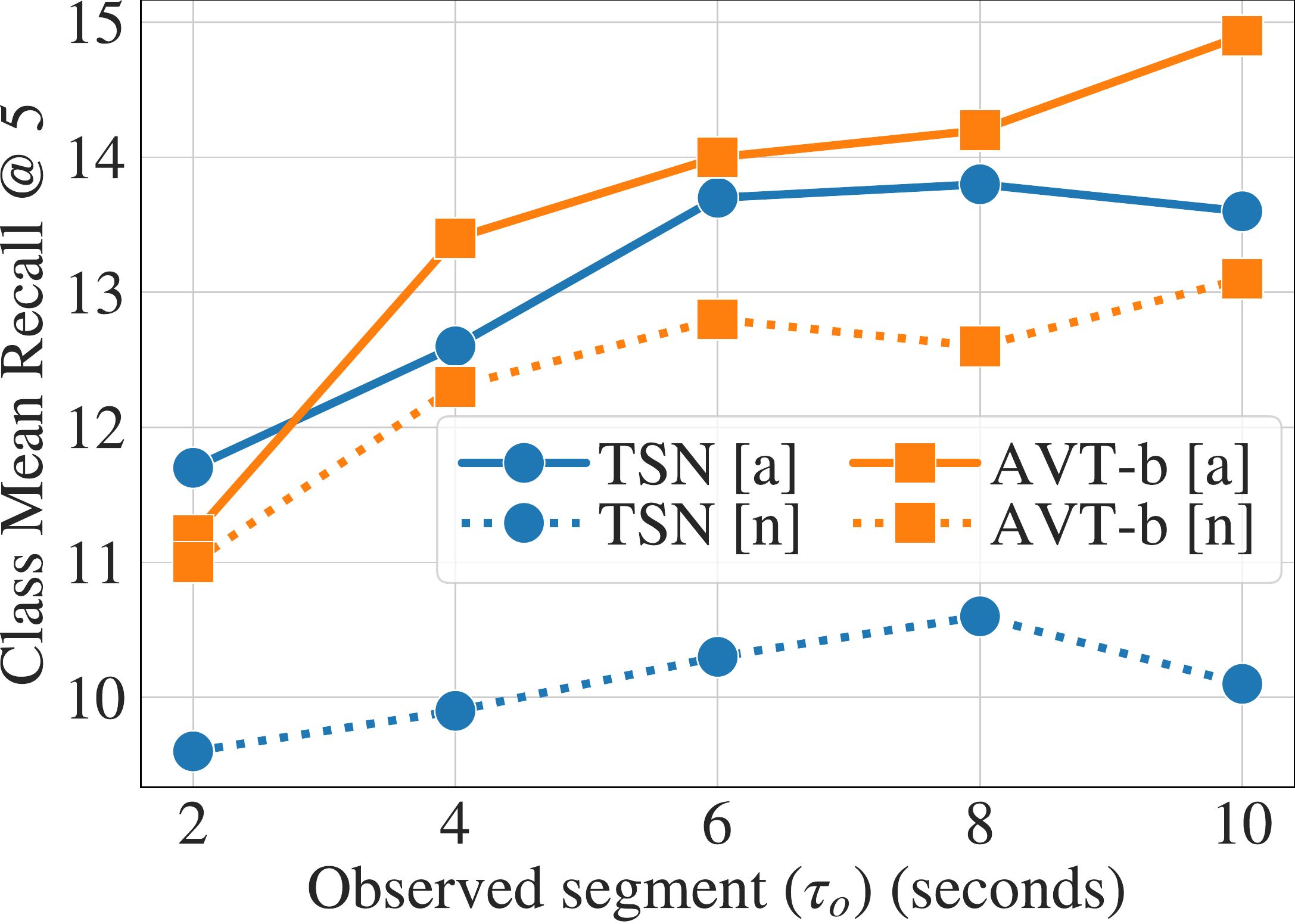}
    \caption{
    {\bf Temporal context.}
    \method effectively leverages longer temporal context, especially in the \causalSettingShort setting.
    }\label{fig:expt:temporal}
\end{minipage}
\end{figure}

\vspace{0.05in}
{\noindent \bf \LossBoth losses.}
In~\cref{tab:abl:losses}, we evaluate the contribution of the two intermediate prediction losses %
that leverage the causal structure of \method.
We find using those objectives leads to significant improvements for both backbones. %
We find $\lossPast$ is more effective for TSN, and $\lossGPT$ for \txBack. Given that both combined work well in both settings, we use both for all experiments. Note that the \acausalSetting setting also serves as a baseline with \txBack backbone followed by simple aggregation on top, and shows our proposed losses encouraging the predictive structure are imperative to obtain strong performance. We analyze per-class gains in~\cref{sec:appdx:per_cls} and find classes like `cook', %
which require understanding the sequence of actions so far to anticipate well, obtain the largest gains in the \causalSetting setting.

\vspace{0.05in}
{\noindent \bf Temporal context.}
Next, we analyze the effect of temporal context. In~\cref{fig:expt:temporal}, we train and test the model with different lengths of temporal context, $\tau_o$. We notice that the performance improves as we incorporate more frames of context, with more consistent gains for \txBack. The gains are especially pronounced when trained using the \causalSetting setting ($11.2\!\rightarrow\!14.9\!=\!3.5\!\uparrow$) %
vs.\ the \acausalSetting ($11.0\!\rightarrow\!13.1\!=\!2.1\!\uparrow$).
This suggests end-to-end trained \method using \lossBoth losses is better suited at modeling sequences of long-range temporal interactions.

{\noindent \bf Attention visualization.}\label{sec:expt:qual:att}
To better understand how \method models videos, we visualize the learned attention in the backbone and head. For the backbone, following prior work~\cite{dosovitskiy2021image}, 
we use attention rollout~\cite{abnar2020quantifying} to 
aggregate attention over heads and layers. For the head, since our causal modeling would bias aggregated attention towards the first few frames, we visualize the last layer attention averaged over heads. As shown in~\cref{fig:teaser}, the model spontaneously learns to attend to hands and objects, which has been found beneficial for egocentric anticipation tasks%
~\cite{liu2020forecasting}---but required manual designation in prior work. The temporal attention also varies between focusing on the past or mostly on the current frame depending on the predicted future action. We show additional results in~\cref{sec:appdx:att_vis}.

\begin{figure}[t]
    \centering
    \includegraphics[width=\linewidth,bb=0 1 908 190]{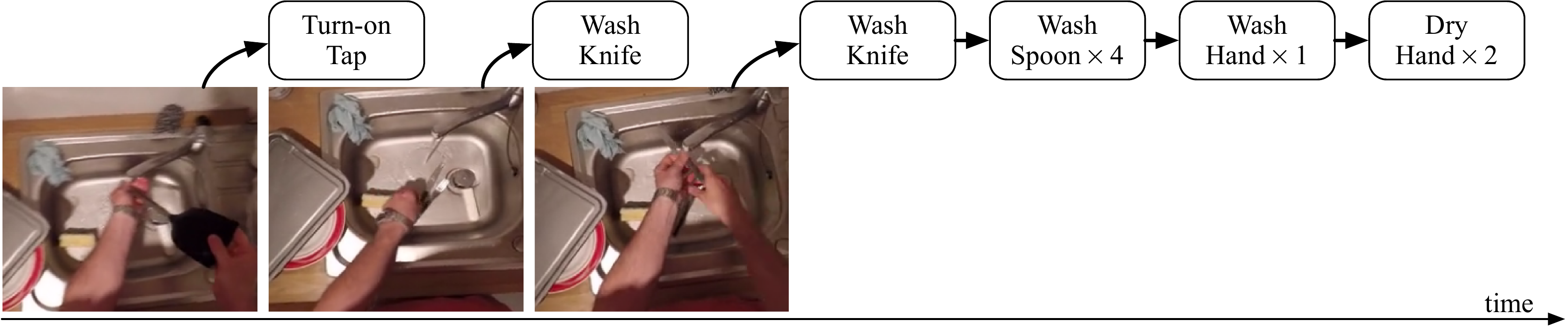}
    \caption{{\bf Long-term anticipation.} \method can also be used to predict further into the future by rolling out predictions autoregressively. 
    The text on top represents the next action predicted at provided frames, 
    followed by subsequently predicted actions, with the number representing how long that action would repeat.
    }\label{fig:expt:dense}
\end{figure}

{\noindent \bf Long-term anticipation.}
So far we have shown \method's applicability in the next-action anticipation task. Thanks to \method's predictive nature, it can also be rolled out autoregressively to predict a sequence of future actions given the video context. We append the predicted feature and run the model on the resulting sequence, reusing features computed for past frames. As shown in~\cref{fig:expt:dense}, \method makes reasonable future predictions---`wash spoon' after `wash knife', followed by `wash hand' and `dry hand'---indicating the model has started to learn %
certain `action schemas'~\cite{piaget1935naissance}, a core capability of our causal attention and \lossBoth training architecture.  We show additional results in~\cref{sec:appdx:dense_anticip}.

\section{Conclusion and Future Work}

We presented \method, an end-to-end attention-based architecture for \lossBoth video modeling. Through extensive experimentation on four popular benchmarks, we show its applicability in anticipating future actions, obtaining \sota results and demonstrating the importance of its \lossBoth training objectives. We believe \method would be a strong candidate for tasks beyond anticipation, such as self-supervised learning~\cite{han2020memdpc,vondrick2016anticipating}, discovering action schemas and boundaries~\cite{shou2021generic,piaget1935naissance}, and even for general action recognition in tasks that require modeling temporal ordering~\cite{goyal2017something}. We plan to explore these directions in future work.

{\noindent \bf Acknowledgements:}
Authors would like to thank Antonino Furnari, Fadime Sener and Miao Liu for help with prior work; Naman Goyal and Myle Ott for help with language models; and Tushar Nagarajan, Gedas Bertasius and Laurens van der Maaten for feedback on the manuscript.

{\small
\bibliographystyle{ieee_fullname}
\bibliography{refs}
}
\clearpage

\appendix

\section{Dataset and Metrics}\label{sec:appdx:dataset}
We test on four datasets as described in the main paper.
{\em \eknewfull (\eknew)}~\cite{damen2020rescaling} is the largest egocentric (first-person) video dataset with 700 long unscripted videos of cooking activities totalling 100 hours.
It contains 89,977 segments labeled with one of 97 verbs, 300 nouns, and 3807 verb-noun combinations (or %
``actions"), and uses $\tau_a$=1s. The dataset is split in 75:10:15 ratio into train/val/test sets, and the test set evaluation requires submission to the CVPR'21 challenge server. 
The evaluation metric used is class-mean recall@5~\cite{furnari2018leveraging}, which evaluates if the correct future class is within the top-5 predictions, and equally weights all classes by averaging the performance computed individually per class. The top-5 criterion also takes into account the multi-modality in the future predictions.
Entries are ranked according to performance on \emph{actions}.   %

{\em \ekfull (\ek)}~\cite{Damen2018EPICKITCHENS} is an earlier version of the \eknew, with 39,596 segments labeled with 125 verbs, 352 nouns, and 2,513 combinations (actions), totalling 55 hours, and  $\tau_a=1s$. We use the standard splits and metrics from~\cite{furnari2020rulstm}.
For anticipation, \cite{furnari2020rulstm} splits the public training set into 23,493 training and 4,979 validation segments from 232 and 40 videos respectively. The test evaluation is similarly performed on the challenge server. The evaluation metrics used are top-1/top-5 accuracies and class-mean recall@5 over verb/noun/action predictions at anticipation time $\tau_a=1s$. 
Unlike \eknew, the recall computation on \ek is done over a subset of `many-shot' classes as defined in~\cite{furnari2019rulstm}.
While \ek is a subset of \eknew, 
we use it to compare to a larger set of baselines, which have not yet been reported on \eknew.

Prior work on the EK datasets~\cite{furnari2019rulstm,sener2020temporal} operate on features from pre-trained models, specifically RGB features extracted using a TSN~\cite{wang2016tsn} architecture trained for action classification on the train set; Flow features using a TSN trained on optical flow; and OBJ features from a Faster R-CNN, whose output is converted into a vector depicting the distribution over object classes for that frame. We refer the reader to~\cite{furnari2019rulstm} for details. We use these features for some experiments in the paper that use fixed backbones (eg TSN). We use the features as provided in the code release by~\cite{furnari2019rulstm}.\footnote{\url{https://github.com/fpv-iplab/rulstm}}

{\em \egtea}~\cite{li2018eye} is another popular egocentric action anticipation dataset, consisting of 10,325 action annotations with 106 unique actions. To be comparable to prior work~\cite{liu2020forecasting}, we report performance on the split 1~\cite{li2018eye} of the dataset at $\tau_a=0.5s$ using overall top-1 accuracy and mean over top-1 class accuracies (class mean accuracy).

Finally, we also experiment with a popular third-person action anticipation dataset:  
{\em \saladfull (\salad)}~\cite{stein2013combining}. It contains fifty $40s$ long videos, with 900 segments labeled with one of 17 action classes. We report top-1 accuracy averaged over the pre-defined 5 splits for an anticipation time $\tau_a=1s$, following prior work~\cite{sener2020temporal,abu2018will}.

\section{Baselines Details}\label{sec:appdx:baselines}
\rulstm leverages a `rolling' LSTM to encode the past, and an `unrolling' LSTM to predict the future, from different points in the past. It was ranked first in the \ek challenge in 2019, and is currently the best reported method on \eknew.
\acbank~\cite{sener2020temporal} improves over \rulstm through a carefully designed architecture leveraging non-local~\cite{wang2017non} and long-term feature aggregation~\cite{wu2019long} blocks over different lengths of past features, and was one of the winners of the CVPR'20 \ek anticipation challenge. Forecasting HOI~\cite{liu2020forecasting} takes an alternate approach, leveraging latest spatio-temporal convnets~\cite{tran2019video} jointly with hand motion and interaction hotspot prediction.

\begin{table}[t]
\centering
\setlength{\tabcolsep}{3pt}
\resizebox{\columnwidth}{!}{%
\begin{tabular}{@{}lccccgc@{}}
\toprule
& \multicolumn{2}{c}{Verb} & \multicolumn{2}{c}{Noun} & \multicolumn{2}{c}{Action} \\
\cmidrule(r){2-3} \cmidrule(lr){4-5} \cmidrule(l){6-7}
Method & \multicolumn{1}{c}{Top-1} & \multicolumn{1}{c}{Top-5} & \multicolumn{1}{c}{Top-1} & \multicolumn{1}{c}{Top-5} & \multicolumn{1}{c}{Top-1} & \multicolumn{1}{c}{Top-5} \\
\midrule
DMR~\cite{vondrick2016anticipating} & - & 73.7 & - & 30.0 & - & 16.9 \\
ATSN~\cite{Damen2018EPICKITCHENS} & - & 77.3 & - & 39.9 & - & 16.3 \\
ED~\cite{gao2017red} & - & 75.5 & - & 43.0 & - & 25.8 \\
MCE~\cite{furnari2018leveraging} & - & 73.4 & - & 38.9 & - & 26.1 \\
FN~\cite{de2018modeling} & - & 74.8 & - & 40.9 & - & 26.3 \\
RL~\cite{ma2016learning} & - & 76.8 & - & 44.5 & - & 29.6 \\
EL~\cite{jain2016recurrent} & - & 75.7 & - & 43.7 & - & 28.6 \\
\fhoi (I3D)~\cite{liu2020forecasting} & 30.7 & 76.5 & 17.4 & 42.6 & 10.4 & 25.5 \\ %
\rulstm~\cite{furnari2020rulstm,furnari2019rulstm} & 32.4 & 79.6 & 23.5 & 51.8 & 15.3 & 35.3 \\  %
\imrnn~\cite{wu2021imaginernn} & - & - & - & - & - & 35.6 \\ %
\acbank~\cite{sener2020temporal} & {\bf 35.8} & {\bf 80.0} & 23.4 & 52.8 & 15.1 & 35.6 \\
\arrayrulecolor{gray}
\midrule
\methodfused & 32.5 & 79.9 & {\bf 24.4} & {\bf 54.0} & {\bf 16.6} & {\bf 37.6} \\ %
\arrayrulecolor{black}
\bottomrule
\end{tabular}}
\caption{{\bf \ek (val) results} reported in top-1/5 (\%) at $\tau_a=1.0s$. The final late-fused model outperforms all prior work.
}
\label{tab:epic55}
\end{table}

\begin{table*}[t]
    \centering
    \setlength{\tabcolsep}{4pt}
    \begin{tabular}{lccgccg|ccgccg}
        \toprule
        & \multicolumn{6}{c|}{Seen test set (S1)} & \multicolumn{6}{c}{Unseen test set (S2)} \\
        & \multicolumn{3}{c}{Top-1 Accuracy\%} & \multicolumn{3}{c|}{Top-5 Accuracy\%} &
        \multicolumn{3}{c}{Top-1 Accuracy\%} & \multicolumn{3}{c}{Top-5 Accuracy\%}
        \\
        \cmidrule(lr){2-4} \cmidrule(lr){5-7} \cmidrule(lr){8-10} \cmidrule(lr){11-13}
        & Verb & Noun & Act. & Verb & Noun & Act.
        & Verb & Noun & Act. & Verb & Noun & Act.
        \\ \midrule
        2SCNN~\cite{Damen2018EPICKITCHENS} & 29.76 & 15.15 & 4.32 & 76.03 & 38.56 & 15.21 & 25.23 & 9.97 & 2.29 & 68.66 & 27.38 & 9.35 \\
        ATSN~\cite{Damen2018EPICKITCHENS} & 31.81 & 16.22 & 6.00 & 76.56 & 42.15 & 28.21 & 25.30 & 10.41 & 2.39 & 68.32 & 29.50 & 6.63 \\
        ED~\cite{gao2017red} & 29.35 & 16.07 & 8.08 & 74.49 & 38.83 & 18.19 & 22.52 & 7.81 & 2.65 & 62.65 & 21.42 & 7.57 \\ %
        MCE~\cite{furnari2018leveraging} & 27.92 & 16.09 & 10.76 & 73.59 & 39.32 & 25.28 & 21.27 & 9.90 & 5.57 & 63.33 & 25.50 & 15.71 \\
        P+D~\cite{miech2019leveraging} & 30.70 & 16.50 & 9.70 & 76.20 & 42.70 & 25.40 & 28.40 & 12.40 & 7.20 & 69.80 & 32.20 & 19.30 \\  %
        \rulstm~\cite{furnari2020rulstm,furnari2019rulstm} & 33.04 & 22.78 & 14.39 & 79.55 & 50.95 & 33.73 & 27.01 & 15.19 & 8.16 & 69.55 & 34.38 & 21.10 \\
        \acbank~\cite{sener2020temporal} & {\bf 37.87} & 24.10 & 16.64 & 79.74 & {\bf 53.98} & 36.06 & 29.50 & 16.52 & 10.04 & 70.13 & 37.83 & 23.42 \\  %
        \fhoi{}~\cite{liu2020forecasting} & 34.99 & 20.86 & 14.04 & 77.05 &  46.45 &  31.29 & 28.27 &  14.07 & 8.64 & 70.67 & 34.35 & 22.91 \\
        \fhoi{}+obj~\cite{liu2020forecasting} & 36.25 & 23.83 & 15.42 & 79.15 & 51.98 & 34.29 & 29.87 & 16.80 & 9.94 & 71.77 & 38.96 & 23.69 \\
        \imrnn~\cite{wu2021imaginernn} & 35.44 & 22.79 & 14.66 & 79.72 & 52.09 & 34.98 & 29.33 & 15.50 & 9.25 & 70.67 & 35.78 & 22.19 \\  %
        Ego-OMG~\cite{dessalene2021forecasting} & 32.20 & {\bf 24.90} & 16.02 & 77.42 & 50.24 & 34.53 & 27.42 & {\bf 17.65} & {\bf 11.81} & 68.59 & 37.93 & 23.76 \\
        \arrayrulecolor{gray}
        \midrule
        \methodfused & 34.36 & 20.16 & {\bf 16.84} & {\bf 80.03} & 51.57 & {\bf 36.52} & {\bf 30.66} & 15.64 & 10.41 & {\bf 72.17} & {\bf 40.76} & {\bf 24.27} \\ %
        \arrayrulecolor{black}
        \bottomrule
    \end{tabular}
    \caption{{\bf \ek test set} results obtained from the challenge server. %
    \method outperforms all published work on this dataset on top-5 metric, and is only second to~\cite{dessalene2021forecasting} on S2 on top-1. Note that~\cite{dessalene2021forecasting} leverages transductive learning (using the test set for initial graph representation learning), whereas \method only uses the train set.
    }
    \label{tab:epic55_test}
\end{table*}

\section{\ekfull Full Results}\label{sec:appdx:ek55_full}

We use the irCSN-152 backbone for comparisons to \sota on \ek, as that performed the best in~\cref{tab:epic55_rgb} on top-1, the primary metric used in \ek leaderboards.
For comparisons using all modalities, in~\cref{tab:epic55}, we late fuse our best model with the other modalities from~\cite{sener2020temporal} (resulting model referred to as \methodfused). We outperform all reported work on the validation set. Finally in~\cref{tab:epic55_test} we train our model on train+val, late fuse other modalities from~\cite{sener2020temporal}, and evaluate on the test sets on the challenge server. Here as well we outperform all prior work. 
Note that our models are only trained for action prediction, and individual verb/noun predictions are obtained by marginalizing over the other.
We outperform all prior work on on seen test set (S1), and are only second to concurrent work~\cite{dessalene2021forecasting} on unseen (S2) for top-1 actions. It is worth noting that~\cite{dessalene2021forecasting} uses transductive learning, leveraging the test set. \method is also capable of similarly leveraging the test data with unsupervised objectives ($\lossGPT$), %
which could potentially further improve in performance. We leave that exploration to future work.
In~\cref{tab:abl:losses_ek55}, we analyze the effect of different losses on the final performance on \ek, similar to the analysis on \eknew in~\cref{tab:abl:losses}. We see a similar trend: using the \causalSetting setting performs the best.

\begin{table} %
    \vspace{0pt}
    \centering
    \setlength{\tabcolsep}{3pt}
    \begin{tabular}{lcccc}
        \toprule
        & \multicolumn{2}{c}{Losses} & \multicolumn{2}{c}{
            Backbones
        } \\
        \cmidrule(r){2-3} \cmidrule(l){4-5}
        Setting & $\lossPast$ & $\lossGPT$ & IG65M & \txBack{} \\ 
        \midrule
        \acausalSetting \acausalSettingShort & - & - & 31.4 & 25.9 \\
        & \cmark  & - & 31.4 & 28.8 \\
        & - & \cmark & {\bf 31.7} &  23.9 \\
        \causalSetting \causalSettingShort & \cmark  & \cmark & {\bf 31.7} & {\bf 30.1} \\
        \bottomrule
    \end{tabular}%
    \captionof{table}{{\bf \LossBoth training on \ek.} Employing the \lossBoth training losses are imperative to obtain strong performance with \method; similar to as seen in~\cref{tab:abl:losses}.
    }\label{tab:abl:losses_ek55}
\end{table}

\section{Analysis} %

\begin{figure}[t]
    \centering
        \includegraphics[width=\linewidth,bb=0 6 1185 283]{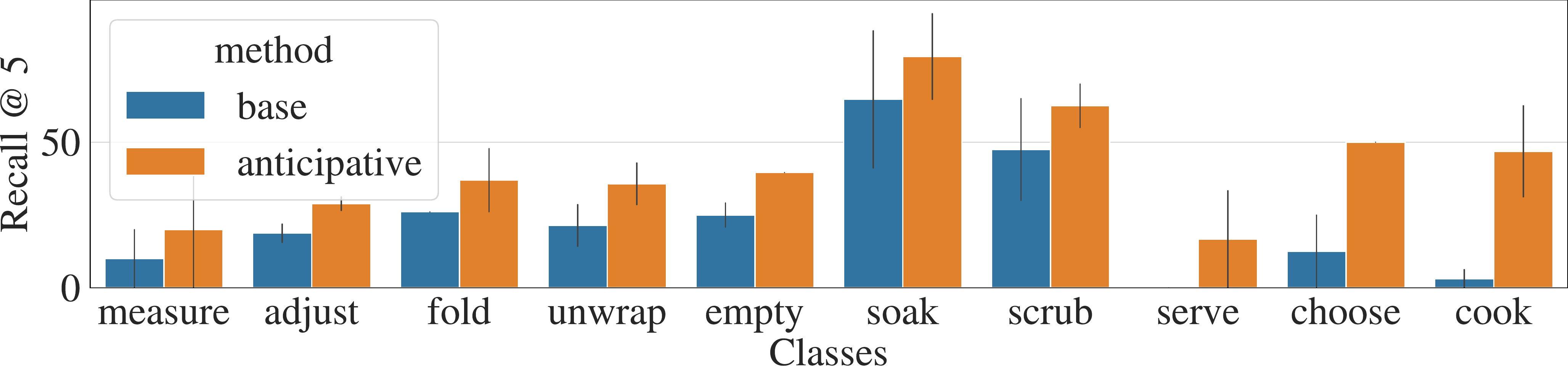}
    \captionof{figure}{{\bf Verb classes that gain the most with causal modeling}, averaged over the TSN and \txBack backbones. Actions such as `cook' and `choose' show particularly significant gains.
    }\label{fig:expt:cls_gains}
\end{figure}

\subsection{Per-class Gains}\label{sec:appdx:per_cls}

To better understand the source of these gains, we analyze the class-level gains with \lossBoth training in~\cref{fig:expt:cls_gains}. We notice certain verb classes show particularly large gains across the backbones, such as `cook' and `choose'. We posit that is because predicting the person will cook an item would often require understanding the sequence of actions so far, such as preparing ingredients, turning on the stove \etc, which the \causalSetting training setting encourages.

\subsection{Attention Visualizations}\label{sec:appdx:att_vis}

In~\cref{fig:appdx:attention} we show additional visualizations of the spatial and temporal attention, similar to~\cref{fig:teaser}. We also show failure cases, which often involve temporal inaccuracy (\ie when the model anticipates an action too soon or too late) and object recognition errors (predicting `wash spoon' instead of `wash fork').
We also provide attached videos to visualize predicted future classes along with the ground truth (GT) future prediction in a video form for \eknew and \egtea, at each time step (as opposed to only 2 shown in these figures).

\subsection{Long-term Anticipation}~\label{sec:appdx:dense_anticip}

In~\cref{fig:appdx:dense_full} we show additional visualizations of the long-term anticipation, similar to~\cref{fig:expt:dense}.

\begin{figure}[t]
    \centering
    \includegraphics[width=0.9\linewidth,bb=0 1 649 615]{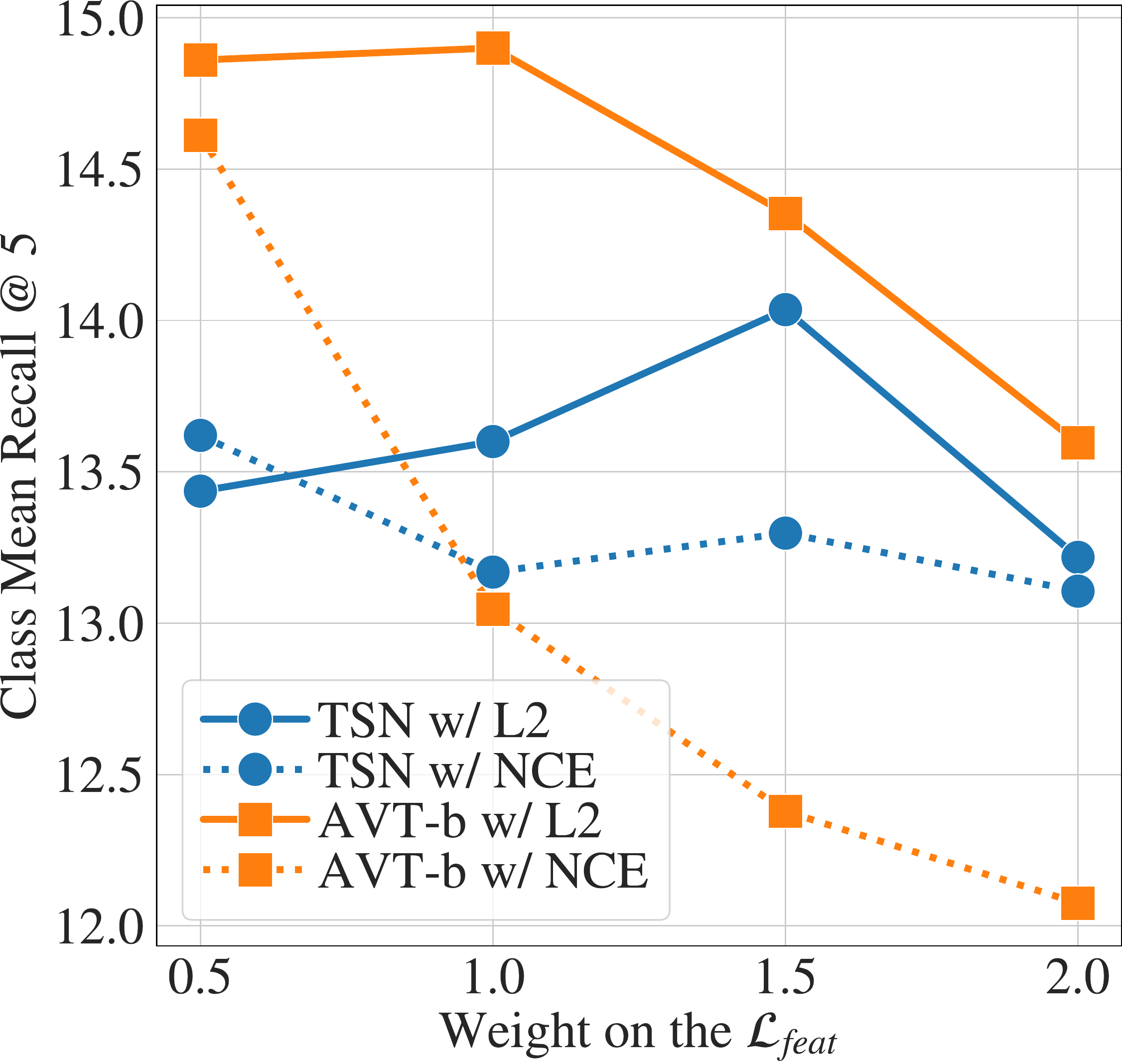}
    \caption{{\bf Different $\lossGPT$ functions and weights}. We found similar or better performance of the simpler $L_2$ metric over NCE and use it for all experiments in the paper. The graph here shows performance on \eknew (validation, RGB) at $\tau_a=1s$, at different scalar weights used on this loss during optimization.
    }\label{fig:appdx:gpt_loss}
\end{figure}

\subsection{$\lossGPT$ Formulation}\label{sec:appdx:gpt_loss}

In~\cref{fig:appdx:gpt_loss} we show the performance of \method with both \txBack and TSN backbones, using two different loss functions for $\lossGPT$: $L_2$ as used in paper, and InfoNCE~\cite{oord2018representation} objective as in some recent work~\cite{han2020memdpc,wu2021imaginernn}, at different weights used on that loss during training. We find that $L_2$ is as effective or better for both backbones, and hence we use it with weight=1.0 for all experiments. While further hyperparameter tuning can potentially lead to further improvements for InfoNCE as observed in some concurrent work~\cite{wu2021imaginernn}, we leave that exploration to future work.

\begin{figure*}[t]
    \centering
    \includegraphics[width=\linewidth,bb=0 0 892 221]{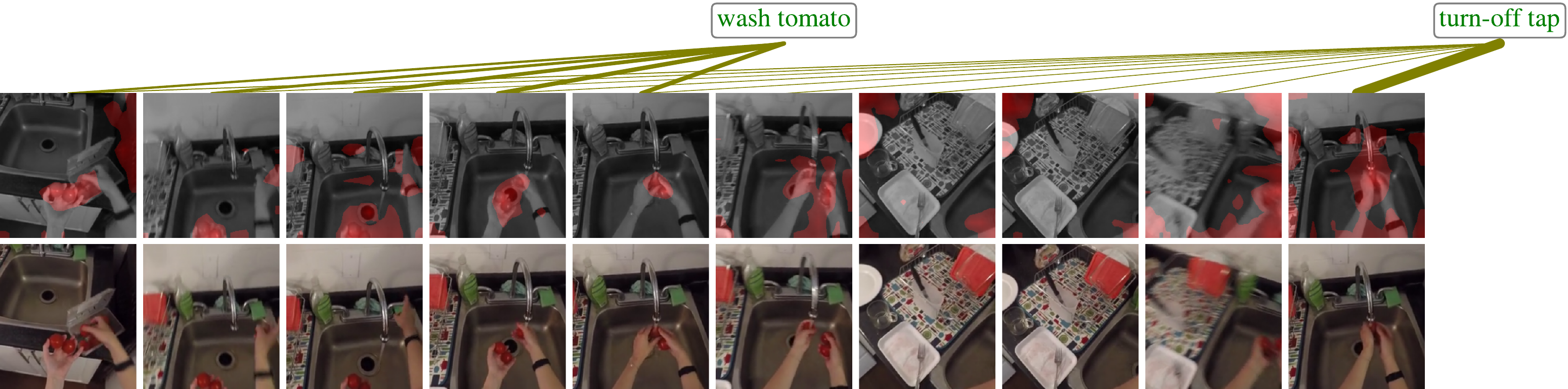}
    \includegraphics[width=\linewidth,bb=0 0 892 221]{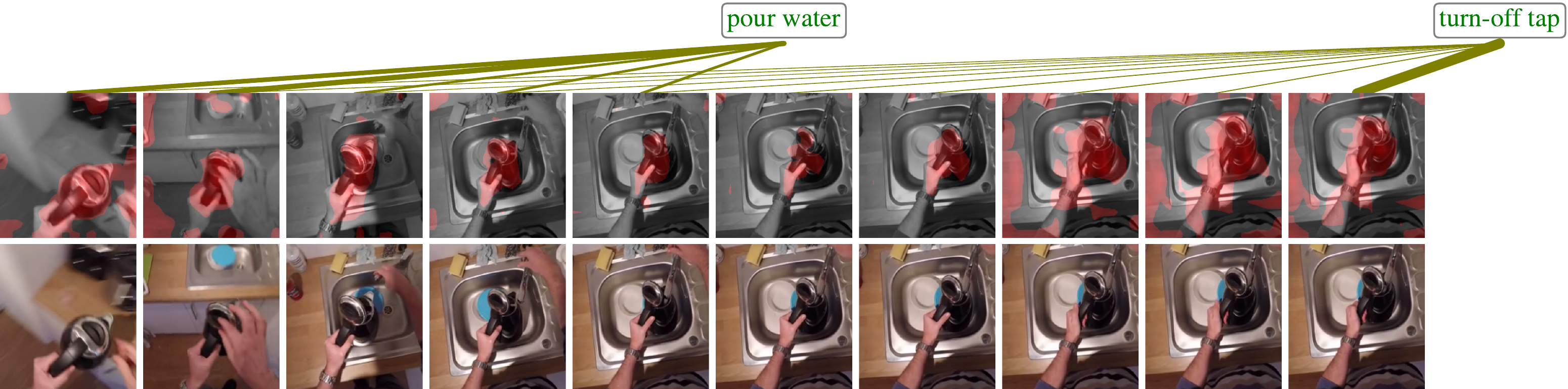}
    \ifshortappdx
    \else
        \includegraphics[width=\linewidth]{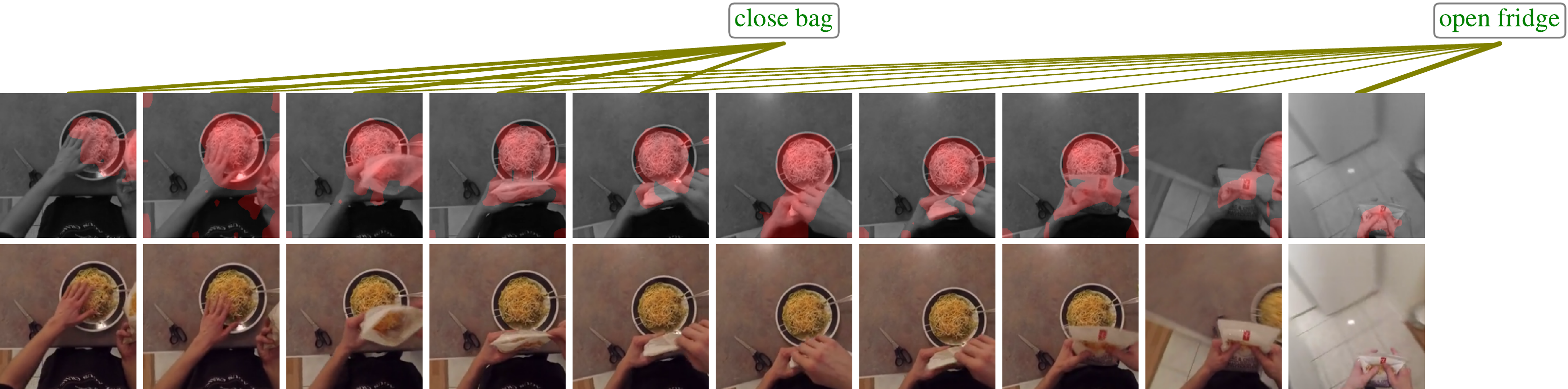}
        \includegraphics[width=\linewidth]{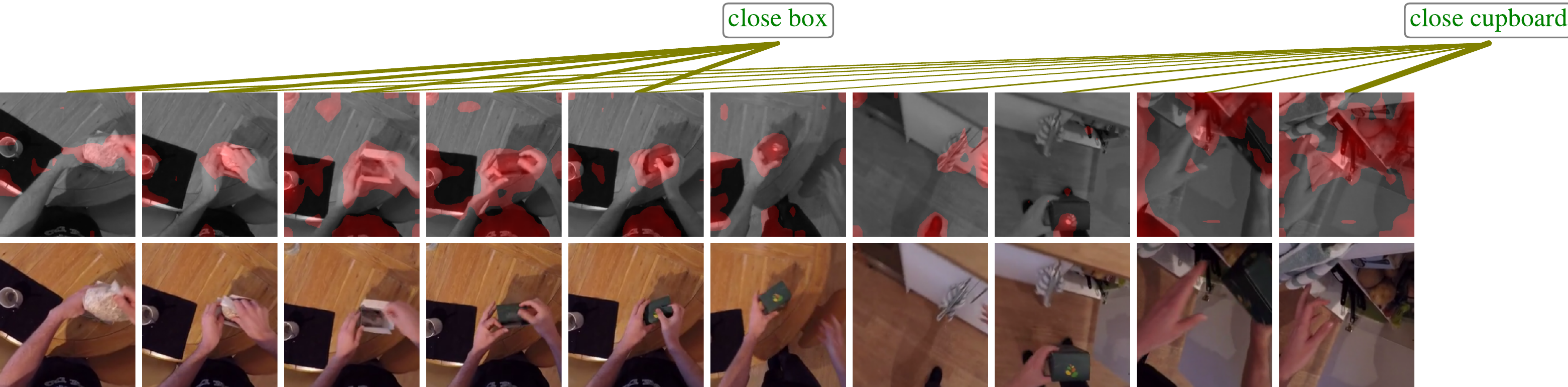}
    \fi
    \caption{{\bf More Qualitative Results.}
    The spatial and temporal attention visualization in \eknew, similar to~\cref{fig:teaser}. For each input frame, we visualize the effective spatial attention by \txBack using attention rollout~\cite{abnar2020quantifying}. The red regions represent the regions of highest attention, which we find to often correspond to hands+objects in the egocentric \eknewfull videos.  The text on the top show future predictions at 2 points in the video, along with the temporal attention (last layer of \txHead averaged over heads) visualized using the width of the lines. The green color of text indicates that it matches the GT action at that future frame (or that nothing is labeled at that frame).  As seen in~\cref{fig:teaser}, spatial attention focuses on hands and objects. The temporal attention focuses on the last frame when predicting actions like `turn-off tap', whereas more uniformly on all frames when predicting `open fridge' (as an action like that usually follows a sequence of actions involving packing up food items and moving towards the fridge).
    \ifshortappdx
        Please refer to the full appendix on the project page for more examples.
    \fi
    }\label{fig:appdx:attention}
\end{figure*}

\ifshortappdx
\else
    \begin{figure*}[t]
        \centering
        \ContinuedFloat 
        \vspace{-0.2in}
        \includegraphics[width=\linewidth]{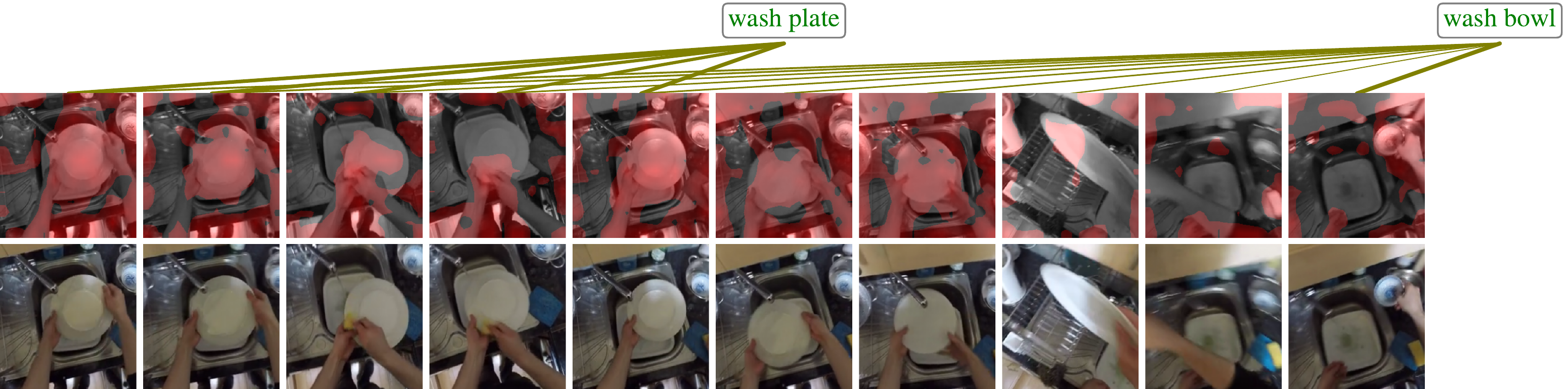}
        \includegraphics[width=\linewidth]{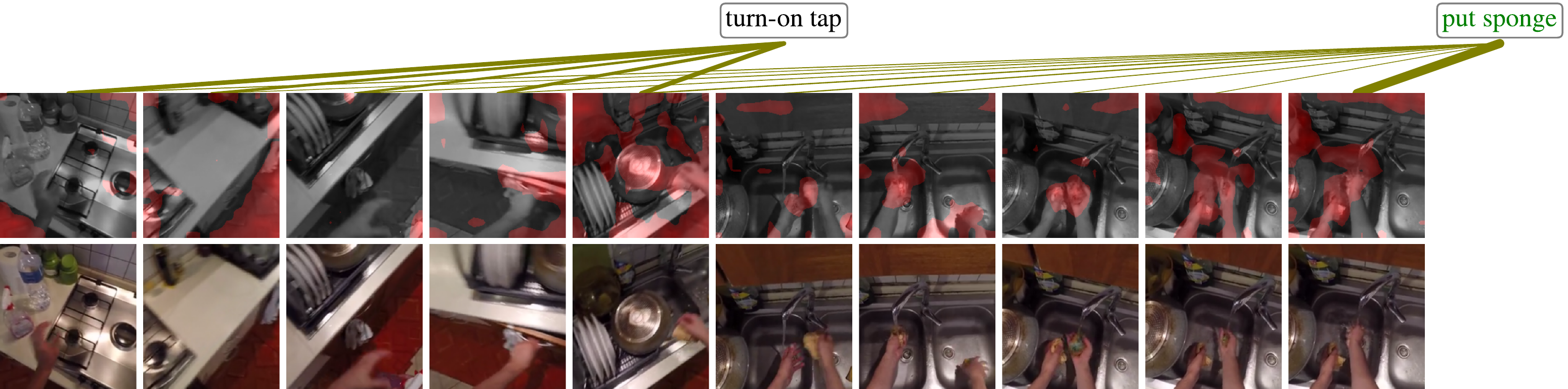}
        \includegraphics[width=\linewidth]{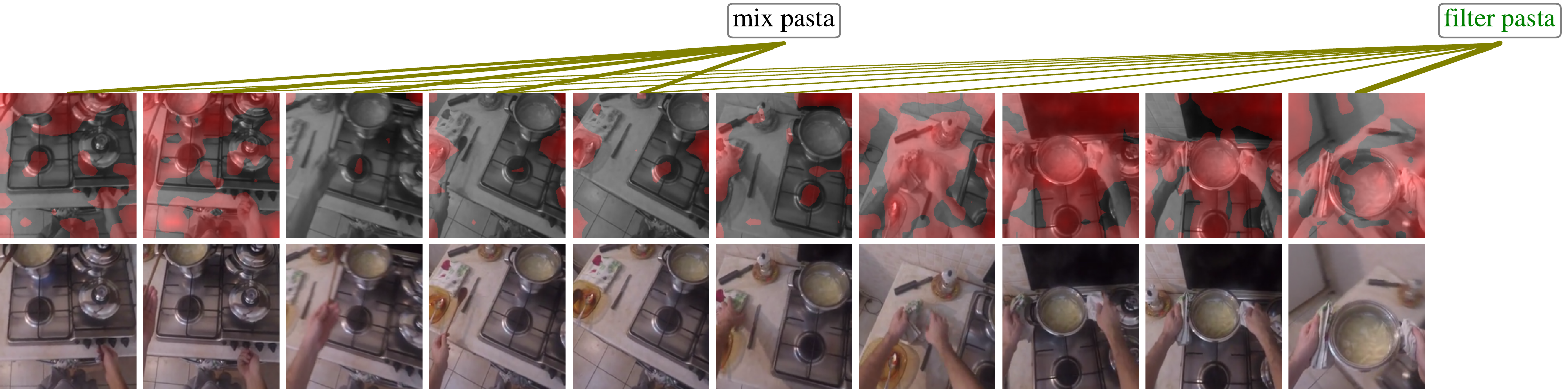}
        \includegraphics[width=\linewidth]{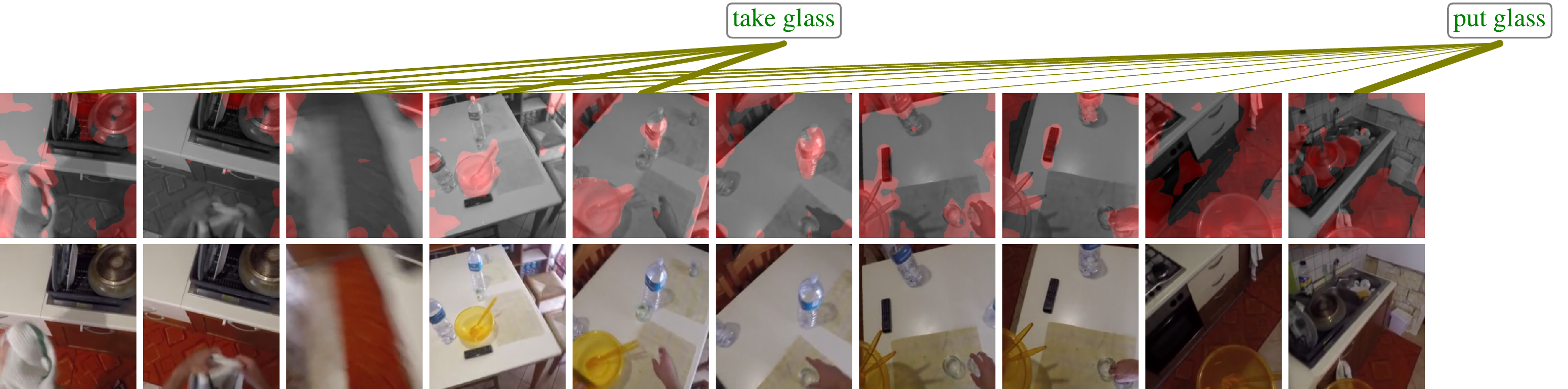}
        \includegraphics[width=\linewidth]{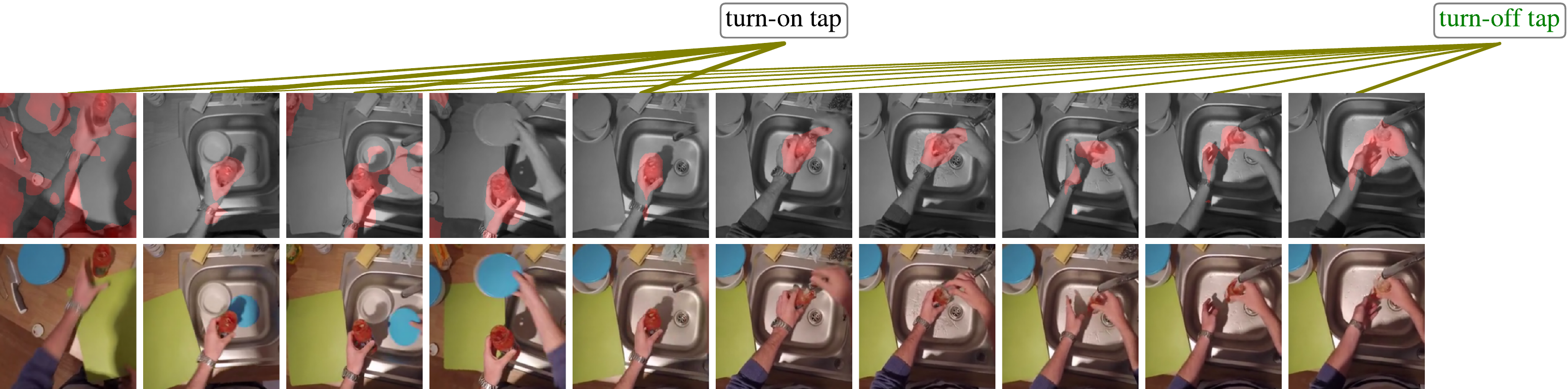}
        \caption{{\bf More Qualitative Results.} (Continued)
        Here we also see some failure cases (the text in black---does not match the labeled ground truth). Note that the predictions in those failure cases are still reasonable. For instance in the second example the model predicts `turn-on tap', while the groundtruth on that frame is `wash cloth'. As we can see in the frame that the water is running, hence the `turn-on tap' does happen before the eventual labeled action of `wash cloth', albeit slightly sooner than when the model predicts.
        }\label{fig:appdx:attention2}
    \end{figure*}
\fi

\begin{figure*}[t]
    \centering
    \includegraphics[width=\linewidth]{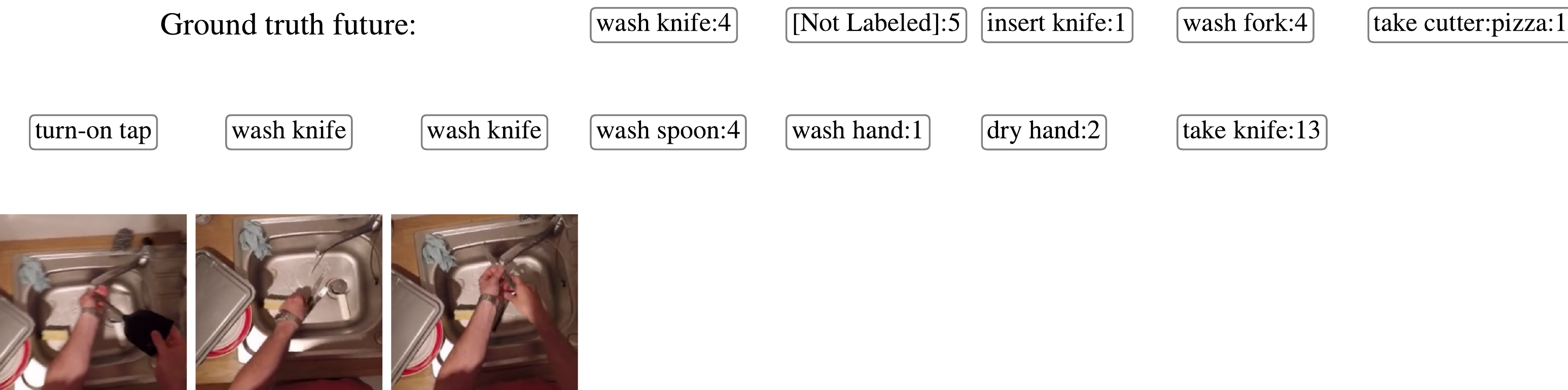}
    \includegraphics[width=\linewidth]{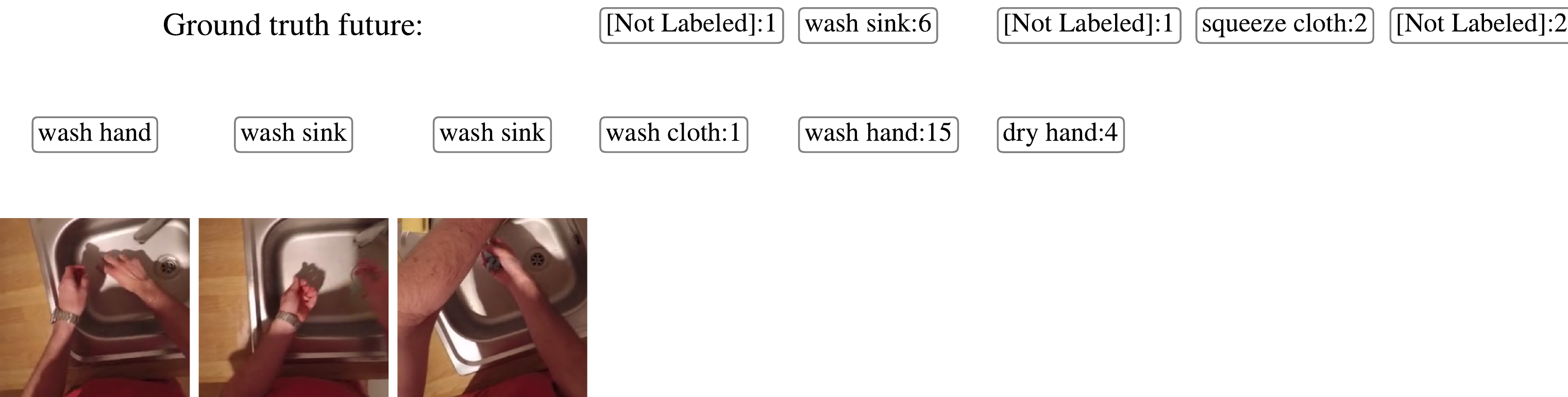}
    \includegraphics[width=\linewidth]{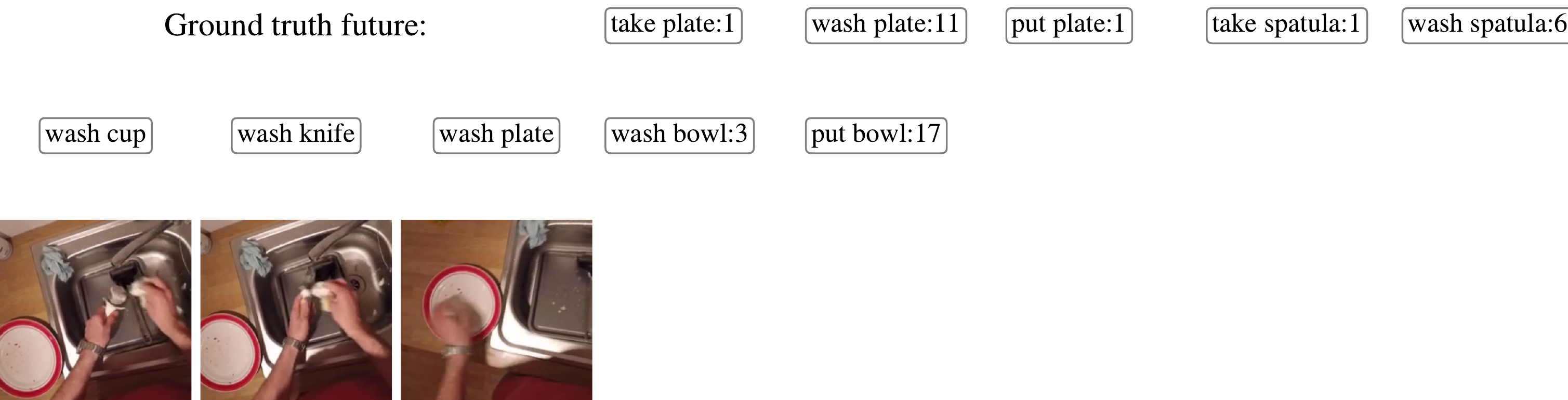}
    \caption{{\bf Long-term anticipation.} Additional results continued from~\cref{fig:expt:dense} on \eknew.
    On top of each frame, we show the {\em future} prediction at that frame (not the action that is happening in the frame, but what the model predicts will happen next). The following text boxes show the future predictions made by the model by rolling out autoregressively, using the predicted future feature. The number next to the rolled out predictions denotes for how many time steps that specific action would repeat, according to the model. For example, `wash spoon: 4' means the model anticipates the `wash spoon' action to continue for next 4 time steps. 
    On top of the predictions we show the labeled ground truth future actions.
    As we can observe, \method makes reasonable future predictions, such as `put pan' would follow `wash pan'; `dry hand' would follow `wash hand' \etc. This suggests the model has picked up on action schemas~\cite{piaget1935naissance}.
    \ifshortappdx
        Please refer to the full appendix on the project page for more examples.
    \fi
    }\label{fig:appdx:dense_full}
\end{figure*}

\ifshortappdx
\else
    \begin{figure*}[t]
        \ContinuedFloat 
        \centering
        \includegraphics[width=\linewidth]{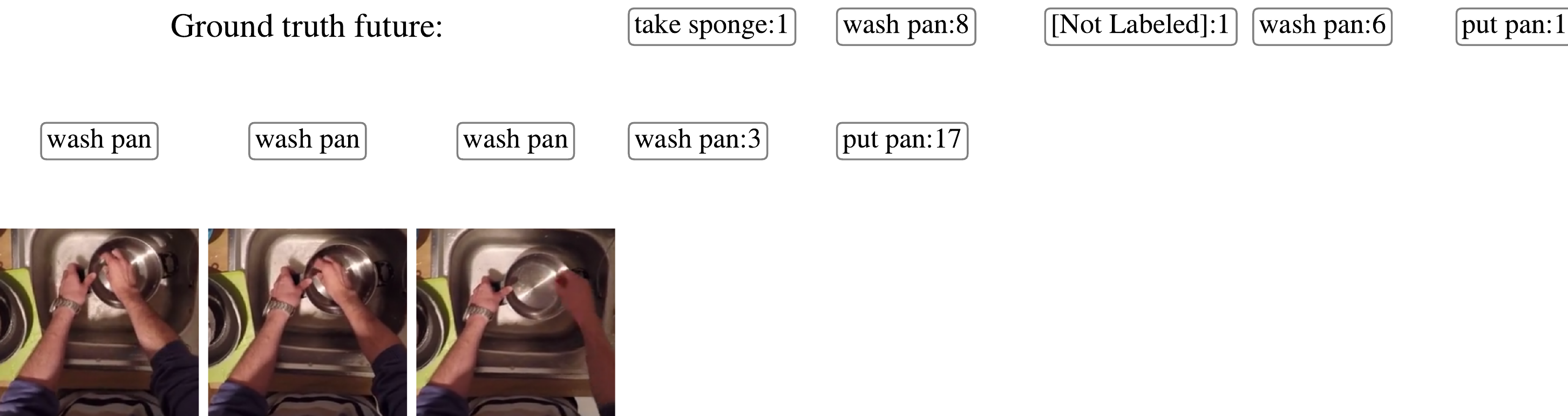}
        \includegraphics[width=\linewidth]{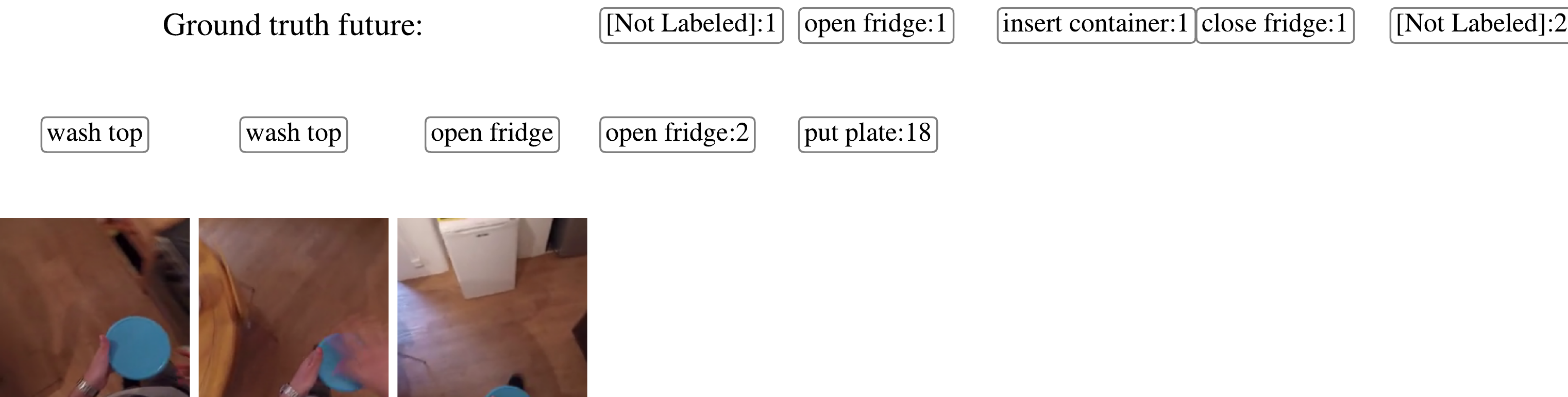}
        \includegraphics[width=\linewidth]{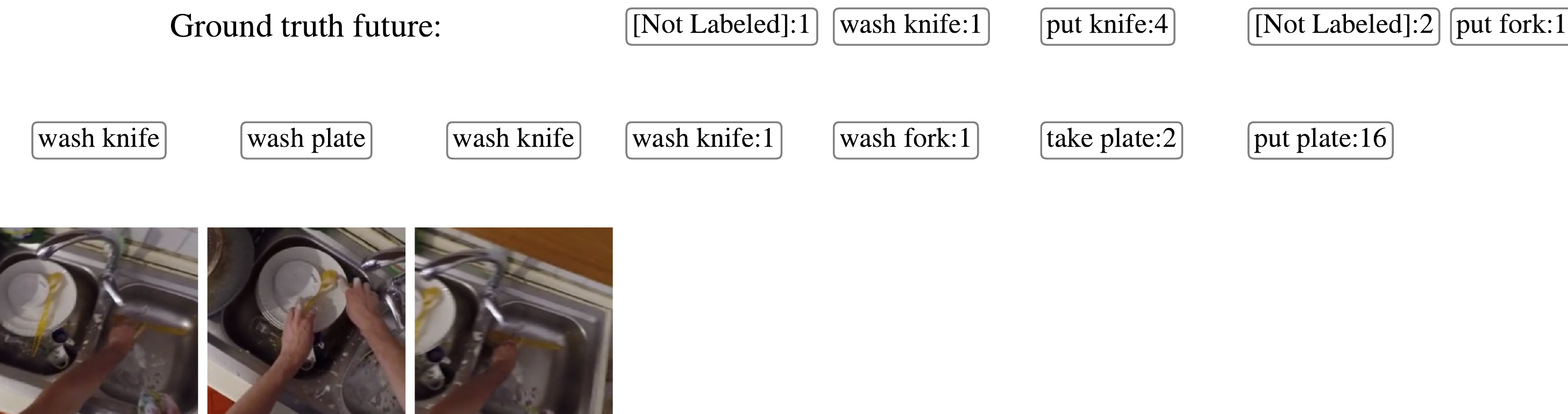}
        \caption{{\bf Long-term anticipation.} (Continued)
        }\label{fig:appdx:dense_full}
    \end{figure*}
\fi

\subsection{Computational complexity}~\label{sec:appdx:complexity}
The only additional compute in \causalSetting training as opposed to \acausalSetting is for applying the linear layer to classify past frame features for $\lossPast$, since $\lossGPT$ simply matches
past features, which anyway need to be computed for self attention to predict the next action. We found GPU memory
remains nearly same, and runtime was only 1\% higher than
a model that only predicts the next action. Also, this additional processing is only for training; inference is exactly
same irrespective of additional losses.

\end{document}